\documentclass[letterpaper,10pt,conference]{ieeeconf}

\IEEEoverridecommandlockouts

\usepackage{amsmath}
\usepackage{amssymb}
\usepackage{adjustbox}
\usepackage{booktabs}
\usepackage{float}
\usepackage{graphicx}
\makeatletter
\let\NAT@parse\undefined
\makeatother
\usepackage[numbers,sort&compress]{natbib}

\usepackage[table]{xcolor}
\usepackage[colorlinks=true,linkcolor=black,citecolor=black,urlcolor=blue]{hyperref}
\usepackage[capitalise]{cleveref}

\definecolor{openlexRow}{HTML}{EAF2F8}
\definecolor{openlexS}{rgb}{0.815,0.935,0.880}
\definecolor{openlexD}{rgb}{0.600,0.820,0.910}
\definecolor{openlexVS}{rgb}{0.940,0.740,0.420}
\definecolor{openlexC}{rgb}{0.860,0.640,0.820}
\definecolor{openlexM}{rgb}{0.820,0.820,0.820}
\definecolor{openlexI}{rgb}{0.955,0.755,0.750}
\definecolor{openlexSHead}{rgb}{0.630,0.870,0.760}
\definecolor{openlexDHead}{rgb}{0.460,0.740,0.870}
\definecolor{openlexVSHead}{rgb}{0.900,0.640,0.260}
\definecolor{openlexCHead}{rgb}{0.760,0.500,0.730}
\definecolor{openlexMHead}{rgb}{0.700,0.700,0.700}
\definecolor{openlexIHead}{rgb}{0.910,0.510,0.500}

\newcommand{\seghead}[2]{\cellcolor{#1}\textbf{#2}}

\def\eqref#1{Eq.~(\ref{#1})}

\def\eg{\emph{e.g.,}}

\usepackage{glossaries-extra}
\setabbreviationstyle[acronym]{long-short}
\glssetcategoryattribute{acronym}{nohyper}{true}

\newacronym{slam}{SLAM}{Simultaneous Localization and Mapping}
\newacronym{ba}{BA}{Bundle Adjustment}
\newacronym{sfm}{SfM}{Structure from Motion}
\newacronym{pgo}{PGO}{Pose-Graph Optimization}
\newacronym{vpr}{VPR}{Visual Place Recognition}
\newacronym{sgd}{SGD}{Stochastic Gradient Descent}
\newacronym{ils}{ILS}{Iterative Least-Squares}
\newacronym{gn}{GN}{Gauss-Newton}
\newacronym{lm}{LM}{Levenberg-Marquardt}
\newacronym{sdp}{SDP}{Semi-Definite Programming}
\newacronym{vo}{VO}{Visual Odometry}
\newacronym{vio}{VIO}{Visual-Inertial Odometry}
\newacronym{imu}{IMU}{Inertial Measurement Units}
\newacronym{pnp}{PnP}{Perspective-n-Point}
\newacronym{dof}{DoF}{Degrees of Freedom}
\newacronym{ar}{AR}{Augmented Reality}
\newacronym{sota}{SOTA}{state-of-the-art}
\newacronym{rpr}{RPR}{Relative Pose Regression}
\newacronym{tsdf}{TSDF}{Truncated Signed Distance Field}
\newacronym{esdf}{ESDF}{Euclidean Signed Distance Field}
\newacronym{gvd}{GVD}{Generalized Voronoi Diagram}
\newacronym{gnc}{GNC}{Graduated Non-Convexity}
\newacronym{gcnn}{GCNN}{Graph Convolutional Neural Network}
\newacronym{llm}{LLM}{Large Language Models}
\newacronym{vlm}{VLMs}{Vision-Language Models}
\newacronym{vl}{VL}{Vision-Language}
\newacronym{ssl}{SSL}{Self-Supervised Learning}
\newacronym{ap}{AP}{Average Precision}
\newacronym{iou}{IoU}{Intersection over Union}
\newacronym{pca}{PCA}{Principal Component Analysis}

\usepackage{etoolbox}
\AtBeginEnvironment{algorithmic}{\everymath{\small}}

\def\slam{\gls{slam} }

\def\tsdf{\gls{tsdf}}

\def\vl{\gls{vl}}
\def\vlm{\gls{vlm}}
\def\ssl{\gls{ssl} }
\def\ap{\gls{ap}}

\definecolor{revisionColor}{HTML}{FF0000}
\definecolor{backcolour}{rgb}{0.95,0.95,0.92}

\usepackage{amsopn}

\newcommand{\bF}{\mathbf{F}}

\newcommand\norm[1]{\left\lVert#1\right\rVert}

\newcommand{\cN}{\mathcal{N}}
\newcommand{\cM}{\mathcal{M}}

\newcommand{\cT}{\mathcal{T}}

\newcommand{\cG}{\mathcal{G}}

\newcommand{\cE}{\mathcal{E}}

\newcommand{\bb}{\mathbf{b}}

\newcommand{\bq}{\mathbf{q}}

\newcommand{\bx}{\mathbf{x}}
\newcommand{\by}{\mathbf{y}}

\newcommand{\bff}{\mathbf{f}}
\newcommand{\bp}{\mathbf{p}}

\def\g2o{$g^2o$}
\def\t2v{\mathrm{t2v}}
\def\v2t{\mathrm{v2t}}
\def\ev2t{\mathrm{ev2t}}

\newcommand{\mypar}[1]{\par\vspace{-1pt} \noindent\textbf{#1}.}

\newcommand{\method}{\textsc{\small{TrackGraph}}}
\newcommand{\methodtitle}{\textsc{TrackGraph}}

\newcommand{\dino}{\mathrm{DINO}}
\newcommand{\ov}{\mathrm{ov}}

\newcommand{\level}{L}

\newcommand{\mesh}{M}
\newcommand{\object}{O}
\newcommand{\segment}{S}

\newcommand{\meshlayer}{\level_\mesh}
\newcommand{\objectlayer}{\level_\object}
\newcommand{\segmentlayer}{\level_\segment}

\newcommand{\point}[2]{\bp_{#1}^{#2}}

\newcommand{\graph}{\cG}
\newcommand{\nodes}[1]{\cN_{#1}}
\newcommand{\edges}[2]{\cE_{#1}^{#2}}
\newcommand{\node}[1]{N_{#1}}

\newcommand{\bbox}[1]{\bb_{#1}}
\newcommand{\feature}[2]{\bff_{#1}^{#2}}
\newcommand{\featurematrix}[2]{\bF_{#1}^{#2}}

\newcommand{\image}[2]{I_{#1}^{#2}}

\title{\LARGE \bf \methodtitle: Online Open-Vocabulary 3D Scene Graphs via Image-Space Tracking}

\author{Peder Borge Hellesylt, Albert Gassol Puigjaner, Kostas Alexis, Annette Stahl
\thanks{Norwegian University of Science and Technology (NTNU), Trondheim, Norway,
    {\tt \footnotesize
        \href{mailto:pederb.hellesylt@gmail.com}{pederb.hellesylt@gmail.com}}
    }
    \thanks{This work was supported in part by the Research Council of Norway under Grant NCEI (No. 357451) and in part by the European Commission under the Horizon Europe Programme through Grant SYNERGISE (No. 101121321).}
}

\definecolor{lightblue}{rgb}{0.12,0.49,0.85}
\usepackage[capitalise]{cleveref}

\usepackage{titlesec}
\titlespacing*{\subsection}{0pt}{1.5ex}{0.5ex}
\begin{document}

\maketitle

\begin{abstract}
Open-vocabulary 3D maps enable robots to reason about previously unknown environments using natural language. However, existing systems typically segment every incoming image, associate detections with persistent 3D segments, and frequently perform costly \vl~inference. We present \method{}, an online open-vocabulary system that maintains short-term 2D mask identity directly in the image stream before fusing segments into 3D. FastSAM masks and CLIP features are computed at sparse keyframes, while dense DINOv3 features are used to propagate masks at a high rate in between. The resulting tracked masks are fused into a class-agnostic 3D segment layer within a hierarchical scene graph, with 3D association handling tracking interruptions and long-term revisits. Compact multi-view CLIP embeddings enable open-vocabulary retrieval.
Across Replica, ScanNet++, and HM3D, \method{} achieves competitive open-vocabulary segmentation and retrieval against state-of-the-art mapping methods, including the highest synonym frequency on Replica ($0.50$). On the same NVIDIA A100, it is $1.7\times$ faster and uses $3.3\times$ less GPU memory than ViT-H OVI-MAP. Real-world quadruped deployments demonstrate onboard scene graph construction and object search at $7.5\mathrm{Hz}$, while recorded drone data is used to test the method under aerial viewpoints.
\end{abstract}

% !TeX root = ../main.tex
\glsresetall
\section{INTRODUCTION}
\label{sec:intro}
Enabling robots to search for objects in environments using natural language requires spatial representations that combine metric geometry with persistent object-level entities and semantic information. Traditional \slam systems estimate the robot trajectory and reconstruct the environment's geometry, but do not organize the reconstructed scene into individual queryable segments. Metric-semantic maps enrich geometry with semantic labels, but may still rely on a predefined vocabulary and do not necessarily preserve object-level identity.

3D scene graphs provide a structure for combining metric geometry with persistent, queryable object-level entities. Hydra~\cite{hughes2022hydra} constructs this hierarchy incrementally, but its object layer depends on closed-set semantic predictions. Open-vocabulary approaches instead associate \vl~features with mapped segments, allowing robots to retrieve segments using language concepts beyond a predefined vocabulary. ConceptGraphs~\cite{Gu2024conceptgraphs} and HOV-SG~\cite{werby2023hovsg} incorporate such segments into scene graphs by lifting frame-local masks into 3D and consolidating observations across views. More recent online mapping methods associate masks through an accumulated volumetric map: OVI-MAP~\cite{deng_ovi-mapopen-vocabulary_2026} votes over instance labels in intersected \tsdf{} voxels, while FindAnything~\cite{laina_findanything_2026} matches masks to mapped segments rendered into the current view. In these approaches, the identity of each frame-local mask is determined from the 3D reconstruction and estimated pose.

\begin{figure}[t]
    \centering
    \includegraphics[
        width=\columnwidth,
        trim=10mm 5mm 17mm 0mm,
        clip
    ]{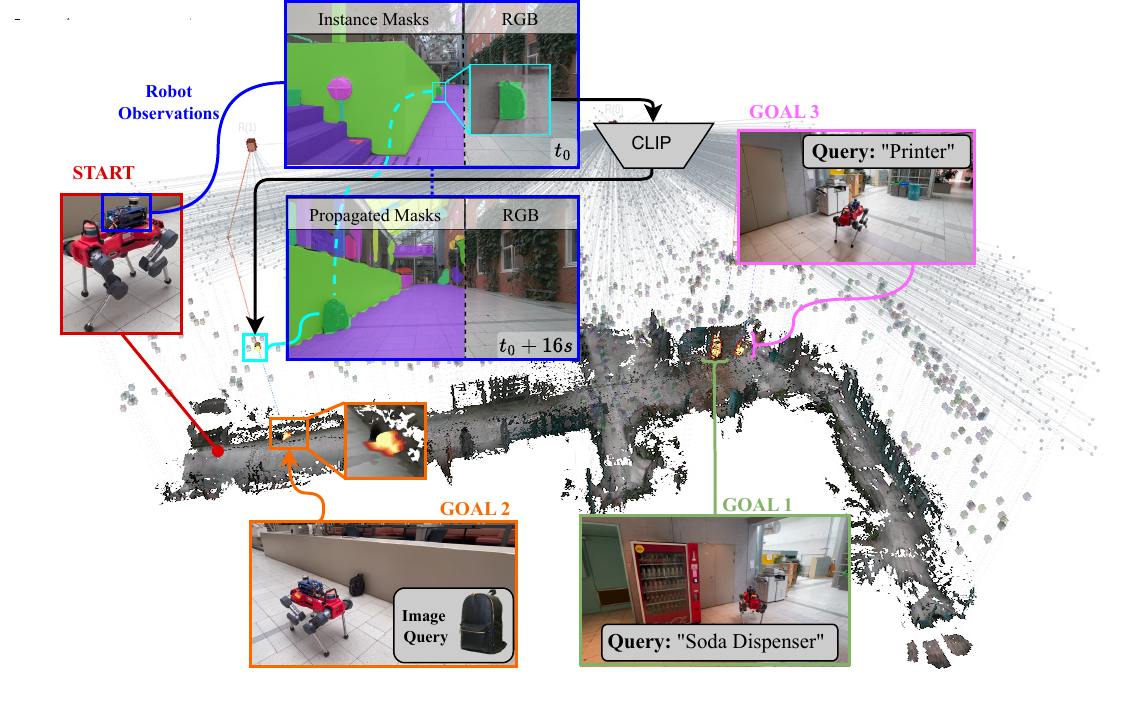}
    \vspace{-2em}
    \caption{\method~constructs an online open-vocabulary 3D scene graph during autonomous robot exploration. 2D source tracks are fused into persistent 3D segments, enabling post-exploration natural-language retrieval and navigation without offline reconstruction or post-processing.}
    \label{fig:intro}
    \vspace{-1ex}
\end{figure}

We introduce \method{}, a method that tracks mask identities in image space before fusing them into a class-agnostic 3D segment layer within a scene graph. FastSAM~\cite{zhao_fast_2023} masks and CLIP~\cite{Radford2021CLIP} features are computed at sparse \emph{keyframes}, while dense DINOv3 features~\cite{simeoni_dinov3_2025} are used to propagate the masks and their identities through the image stream in between. We refer to each 2D temporally tracked mask identity as a \emph{source track}. Source track masks are continually fused into a \tsdf{} and mesh, where repeated observations reinforce consistent track identities that are then used to build persistent 3D segments. Interrupted tracks and long-term revisits are reconciled via 3D geometric and visual association, while CLIP features~\cite{Radford2021CLIP} are averaged per source track and stored as separate views in each 3D segment’s feature gallery. As shown in~\cref{fig:intro}, \method{} incrementally constructs the scene graph during exploration, enabling language- and image-guided object search. 

Our main contributions are:

\begin{itemize}
    \item We introduce an image-space tracker that propagates sparse FastSAM masks with dense DINOv3 features, preserving short-term identity without per-frame segmentation and \vl~inference.
    \item We develop an online open-vocabulary 3D segment map within a hierarchical scene graph by fusing source tracks into a \tsdf{}, constructing persistent class-agnostic segments, reconciling tracking interruptions and long-term revisits, and retaining multi-view CLIP galleries.
    \item We demonstrate competitive open-vocabulary segmentation and retrieval with efficient operation, including low GPU memory use and online onboard deployment. Quadruped experiments demonstrate online scene graph construction and object search, while recorded drone data is used to evaluate mapping and retrieval under distinct aerial viewpoints.
    
\end{itemize}

% !TeX root = ../main.tex

\section{RELATED WORK}\label{sec:related}

\method{}~addresses online open-vocabulary 3D segmentation and scene graph construction using vision foundation models and image-space temporal segment tracking. Therefore, we first review visual features for open-vocabulary representations and mask propagation, and then discuss existing open-vocabulary 3D mapping approaches.

\mypar{Vision Foundation Models}
\vlm~such as CLIP~\cite{Radford2021CLIP} learn aligned image and text embeddings through contrastive pretraining on large-scale internet data, enabling image content to be queried with natural language. CLIP represents an image with an embedding and is commonly used together with class-agnostic segmentation models such as SAM~\cite{Kirillov2023SAM} to obtain open-vocabulary masks. \ssl visual foundation models like DINOv3~\cite{simeoni_dinov3_2025} instead learn dense representations directly from images without language alignment, capturing local visual structure that remains consistent across similar views. 

\mypar{3D Object Maps and Scene Graphs} Open-vocabulary 3D maps differ in where they store \vl~features and how they associate observations across views. Dense methods store features per point~\cite{Peng2023OpenScene,Jatavallabhula2023conceptfusion}, making \vl~storage scale with map size and resolution. HOV-SG~\cite{werby2023hovsg} clusters point-level features within segments to select representative descriptors for a floor-room-segment graph, while ConceptGraphs~\cite{Gu2024conceptgraphs} instead aggregates features directly per 3D segment. Both back-project masks and merge 3D observations using a combination of geometric and semantic overlap. OpenTrack3D~\cite{zhou_opentrack3d_2025} similarly associates lifted masks in 3D using voxel overlap and visual similarity, then refines the resulting 3D proposals before multimodal language model classification.

Recent online open-vocabulary mapping systems increasingly associate or track segments using voxel-based instance maps. OVI-MAP~\cite{deng_ovi-mapopen-vocabulary_2026} votes over instance labels in \tsdf~voxels containing lifted mask points, while OpenVox~\cite{deng_openvox_2025} combines voxel-level instance probabilities with caption-embedding similarity. ThinkGraphs~\cite{bickici2026thinkgraphs} adopts this probabilistic formulation and adds asynchronous VLM agents for merging and graph construction. Other methods perform association by projecting the map into the current image: OVO-SLAM~\cite{martins2024ovoslam} assigns masks by voting over projected point labels, while Open-Fusion~\cite{yamazaki_open-fusion_2023} and FindAnything~\cite{laina_findanything_2026} use overlap between detected and rendered regions.

These approaches typically segment each frame and associate the resulting masks with the 3D map. In contrast, \method{} maintains short-term mask identity directly in image space by propagating FastSAM~\cite{zhao_fast_2023} masks between sparse keyframes using DINOv3~\cite{simeoni_dinov3_2025} features. Consecutive observations can therefore be fused into the \tsdf{} with consistent identities, while 3D association is only needed after tracking interruptions or revisits. Rather than inferring \vl~features at every frame~\cite{Gu2024conceptgraphs, werby2023hovsg, laina_findanything_2026}, we compute CLIP~\cite{Radford2021CLIP} features only at keyframes and aggregate them per source track, reducing repeated segmentation and \vl~inference.

% !TeX root = ../main.tex

\section{PROBLEM STATEMENT}\label{sec:problem}
Given a sequence of $M$ RGB-D observations and corresponding odometry estimates $\mathcal{I}_{1:M}=\{\image{t}{RGB},\,\image{t}{D},\,{X}_t\}_{t=1}^{M}$, the goal is to incrementally construct an open-vocabulary 3D segment map, represented within a hierarchical scene graph $\graph=(\cN,\cE)$. Its node set $\cN$ is organized into layers ranging from dense metric-semantic geometry and persistent 3D segments to higher-level spatial regions, while $\cE$ encodes the relationships within and between these layers. Each 3D segment stores open-vocabulary features, enabling the map to be queried using text or images. The system must incrementally construct the graph online while preserving segment identities across consecutive observations and revisits.

% !TeX root = ../main.tex

\section{METHOD}\label{sec:method}

\begin{figure*}[t]
    \centering
    \includegraphics[
        width=\textwidth,
        trim=0pt 0pt 0pt 0pt,
        clip
    ]{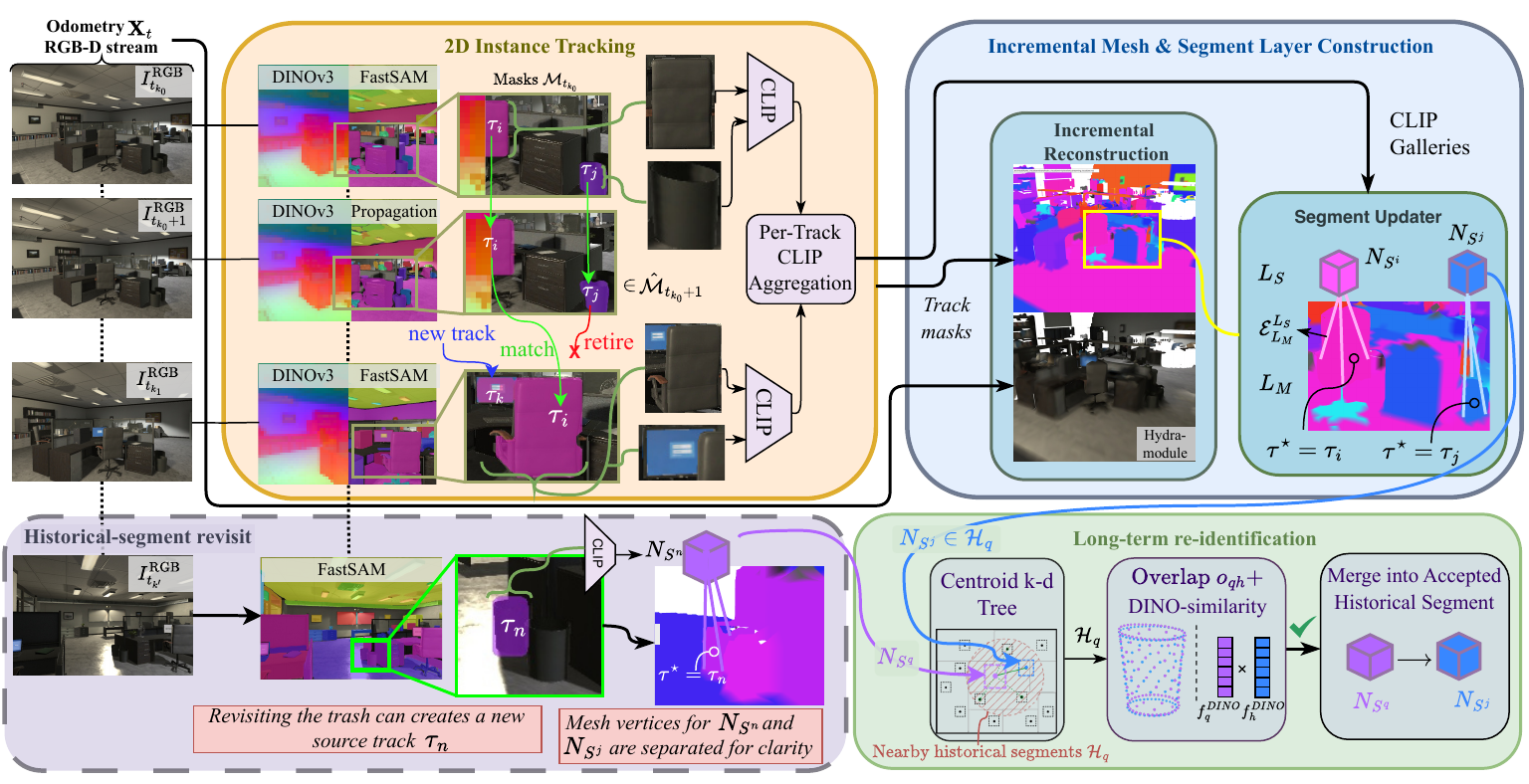}
        \vspace{-2em}
    \caption{\method{} 2D instance tracking, 3D segment construction and long-term re-identification. The upper path propagates sparse FastSAM~\cite{zhao_fast_2023} masks between keyframes using DINOv3~\cite{simeoni_dinov3_2025} features, with feature maps  $\featurematrix{t}{\dino}$ demonstrating how object structures emerge in feature space. The resulting source track masks are continually fused into the mesh from which persistent 3D segments are constructed. CLIP~\cite{Radford2021CLIP} features are aggregated per track for open-vocabulary retrieval. The lower path handles long-term revisits by retrieving nearby historical segments and merging with the most similar.}
    \label{fig:trackgraph-main-figure}
    \vspace{-2em}
\end{figure*}

\method{}, presented in~\cref{fig:trackgraph-main-figure}, generates a hierarchical scene graph with a class-agnostic 3D segment layer driven by image-space mask tracking. FastSAM~\cite{zhao_fast_2023} class-agnostic masks and CLIP~\cite{Radford2021CLIP} features are computed at sparse keyframes, while DINOv3~\cite{simeoni_dinov3_2025} features are used to propagate masks and their identities between keyframes. We call each temporally maintained 2D identity a \emph{source track}. At subsequent keyframes, matched masks are assigned to existing source tracks, unmatched masks initialize new tracks, and unmatched active tracks are retired. Source track masks are fused into the \tsdf{} and mesh to construct persistent 3D segments, while 3D association mechanisms consolidate compatible tracks after interruptions and revisits. By resolving short-term continuity before 3D fusion, we avoid treating every frame-local mask as a 3D association problem.

Section~\ref{subsec:scene-graph-representation} defines the segment-centric graph representation, while \cref{subsec:instance-tracking} describes the instance-tracking frontend. \cref{subsec:source_track_fusion,subsec:segment-construction} present source track fusion and 3D segment construction, respectively. \cref{subsec:segment-association} describes long-term re-identification, and \cref{subsec:ov-querying} presents open-vocabulary feature storage and retrieval.

\subsection{Scene Graph Representation}
\label{subsec:scene-graph-representation}
We use Hydra~\cite{hughes2022hydra} as the online reconstruction and scene graph backbone. Hydra represents the environment as a hierarchical scene graph $\graph = (\cN, \cE)$ in which the mesh layer $\meshlayer$ stores metric-semantic geometry, its object layer $\objectlayer$ contains persistent object nodes, and its higher layers represent places $L_P$, rooms $L_R$, and the building $L_B$. We retain Hydra's graph structure, but replace the object layer $\objectlayer$ with a class-agnostic segment layer $\segmentlayer$.

We modify the mesh layer $\meshlayer$ to store per-vertex source track likelihoods rather than closed-set semantics, supporting the construction of nodes in our segment layer $\segmentlayer$. Each mesh vertex is represented by a node $N_{M} = \{ \point{}{M},\, \mathbf{c},\, H \} \in \mathcal{N}_{L_M}$, where $\point{}{M}\in\mathbb{R}^3$ is its 3D position, $\mathbf{c}$ is its RGB color, and $H_{} = \{(\tau_k,\ell_{\tau_k})\}_{k=1}^{K} $ contains at most $K$ source track hypotheses, each pairing a track identity $\tau_k$ with its accumulated weight $\ell_{\tau_k}$. Mesh nodes may be connected to 3D segment nodes through edges in $\edges{\meshlayer}{\segmentlayer}$.

We define a 3D segment node as $\node{\segment}^{} = \left\{ \bbox{},\, \point{}{S},\, T,\, \bff_{}^{\dino},\, \bF_{}^{\ov},\, s \right\} \in \nodes{\segmentlayer}$, where $\bbox{}$ is its 3D bounding box, $\point{}{S}$ is its centroid, $T$ is the set of source tracks assigned to the segment, $\bff_{}^{\dino}$ is its aggregated DINOv3 feature, $\bF_{}^{\ov}$ is its gallery of CLIP features, and $s$ denotes whether the segment is active or historical. The \tsdf{} and mesh are reconstructed within a spatial \emph{active window} of $8$\,m around the current pose, and a segment is active while at least one of its connected mesh vertices lies within this window and becomes historical once all its vertices have left it. It may also become active again after long-term segment re-identification (\cref{subsec:segment-association}). Each segment node is connected to its nearest place node, and place nodes are connected to rooms.

\subsection{Instance Tracking}
\label{subsec:instance-tracking}
Instance tracking (top-left of~\cref{fig:trackgraph-main-figure}) converts frame-local masks into \emph{source tracks}, temporally persistent 2D mask identities maintained through frame-to-frame propagation and association with subsequent FastSAM masks.

For every processed frame $\image{t}{RGB}$ at time $t$, we extract a dense DINOv3 feature map $\featurematrix{t}{\dino}$ using a ViT-S+ backbone~\cite{simeoni_dinov3_2025}. We index keyframes by $k$, with $t_k$ denoting the corresponding keyframe index. At each keyframe, separated by $K_f$ frames, we additionally run the class-agnostic segmenter FastSAM~\cite{zhao_fast_2023} to obtain $\cM_{t_k}=\{m_{t_k,j}\}_{j=1}^{n_{t_k}}$, where $m_{t_k,j}$ is the $j$-th detected mask at keyframe $k$ and $n_{t_k}$ is the number of detected masks. 

\mypar{Source Track Representation}
At frame index $t$, the tracker maintains the active source track set $\cT_t=\{\tau_i\}_{i=1}^{N_t}$, where $N_t$ is the number of active tracks. A source track is represented as $\tau_i =( \mathrm{id}_i, \bar{\bff}_{i}^{\dino}, \bff_{i}^{\ov}, n^{\ov}_i)$, where $\mathrm{id}_i$ is its persistent identifier,  $\bar{\bff}_{i}^{\dino}$ is its aggregated DINOv3 feature, $\bff_{i}^{\ov}$ is its aggregated open-vocabulary feature, and $n_i^{\ov}$ counts the CLIP features accumulated by the track. Aggregated features are updated as new FastSAM masks are associated with the track, while unmatched active tracks are retired from $\cT_t$.

\mypar{Propagation Using DINOv3 Features}
Between keyframes, we extend the probabilistic label-propagation method introduced by DINOv3~\cite{simeoni_dinov3_2025} to obtain intermediate masks while preserving their source track identities. Let $\mathbf{Y}_t(p)$ denote the probability distribution over active source track identities at DINOv3 patch $p$ in frame $t$.
At keyframe $k$, after each FastSAM mask $m_{t_k,j}\in\cM_{t_k}$ is associated with a source track, the masks initialize $\mathbf{Y}_{t_k}$ with one-hot assignments (\eg~ $\mathbf{Y}_{t_k}(p)=[0,1,0,0]^\top$ for $\tau_2$ at $p$ among four tracks).

For each frame $\image{t}{RGB}$ between consecutive keyframes $\image{t_k}{RGB}$ and $\image{t_{k+1}}{RGB}$, we propagate source track identity from similar patches in earlier frames. For each patch $p$ in frame $\image{t}{RGB}$, we search the preceding keyframe $\image{t_k}{RGB}$ and several recent frames $t_k<r<t$ for reference patches $q$ within a local neighborhood of $p$. We retain the references with highest DINOv3 cosine similarity, normalize their similarities with a softmax, and combine their source track distributions:
\begin{equation}
    \mathbf{Y}_t(p)
    =
    \sum_{(q,r)\in\mathcal{K}_t(p)}
        a_{pqr}\mathbf{Y}_r(q),
    \label{eq:source_track_propagation}
\end{equation}
where $\mathcal{K}_t(p)$ is the set of selected reference patch-frame pairs $(q,r)$ for patch $p$, and $a_{pqr}$ is the corresponding softmax-normalized DINOv3 similarity. The resulting tensor is retained for the next frame and upsampled to image resolution, where a pixel-wise argmax produces the predicted masks $\hat{\cM_{t}}= \{\hat m_{t,i}\}_{\tau_i\in\cT_t}$. Propagation carries existing identities; tracks are initialized, matched, and retired at keyframes.

\mypar{Keyframe Correction} 
The first keyframe initializes the source tracks. At every subsequent keyframe $\image{t_k}{RGB}$, \cref{eq:source_track_propagation} first predicts $\hat \cM_{t_k}$, after which every FastSAM mask $m_{t_k,j}\in\cM_{t_k}$ is compared with every propagated source track mask $\hat m_{t_k, i}$, using overlap and visual similarity. Let $(u,v)$ describe the coordinates of patch $p$ in the DINOv3 feature map $\featurematrix{t_k}{\dino}$.  For each detected mask $m_{t_k,j}$, we average its normalized patch features,
\begin{equation}
    \tilde{\bff}_{t_k,j}^{\dino}
    =
    \frac{1}{|m_{t_k,j}|}
    \sum_{(u,v)\in m_{t_k,j}}
    \frac{
        \featurematrix{t_k}{\dino}(u,v)
    }{
        \left\|\featurematrix{t_k}{\dino}(u,v)\right\|_2
    },
    \label{eq:mask-prototype}
\end{equation}
and normalize the result as $\bff_{t_k,j}^{\dino}
=
\tilde{\bff}_{t_k,j}^{\dino}/
\|\tilde{\bff}_{t_k,j}^{\dino}\|_2$, obtaining a single visual descriptor for each FastSAM mask.
We score the association between an active track's $\tau_i$ predicted mask $\hat m_{t_k,i}$ and a detected mask $m_{t_k,j}$ by
\begin{equation}
\begin{aligned}
    s_{ij}
    &=
    w_{\mathrm{IoU}}\,
    \mathrm{IoU}(\hat m_{t_k, i},m_{t_k,j})
    \\
    &\quad+
    w_{\mathrm{feat}}\,
    \max\!\left(
        \cos\!\left(
            \bar{\bff}_{i}^{\dino},
            \bff_{t_k,j}^{\dino}
        \right),
        0
    \right),
\end{aligned}
\label{eq:keyframe_assoc}
\end{equation}
where $w_{\mathrm{IoU}}+w_{\mathrm{feat}}=1$, and IoU denotes Intersection over Union. Candidate pairs must meet minimum thresholds on both $\mathrm{IoU}$ and the combined association score $s_{ij}$, which prevent association based on appearance without spatial overlap, or on overlap between visually dissimilar masks.

Valid candidates are resolved by greedy one-to-one assignment, where matched detections continue existing tracks, unmatched detections initialize new tracks, and unmatched active tracks are retired and no longer propagated. We retire tracks when one cannot be continued reliably, for example when an object leaves the camera view or is fragmented substantially. For every accepted pair $(i,j)$, the source track DINOv3 feature is updated as $\bar{\bff}_{i}^{\dino}\leftarrow(1-\eta)\bar{\bff}_{i}^{\dino}+\eta\bff_{t_k,j}^{\dino}$, adapting it to recent observations while retaining the track's appearance history.

\mypar{Open-vocabulary Features}
For every keyframe mask $m_{t_k,j}$, we extract a padded bounding-box crop and compute its CLIP~\cite{Radford2021CLIP} feature $\bff_{t_k,j}^{\ov}$. We compute CLIP features asynchronously to avoid delaying propagation. Their use is thus reserved for downstream retrieval (\cref{subsec:ov-querying}). 

Similarly to FindAnything~\cite{laina_findanything_2026}, when $m_{t_k,j}$ is assigned to source track $\tau_i$, its open-vocabulary feature is fused into the track average,
\begin{equation}
    \bff_{i}^{\ov}
    \leftarrow
    \frac{
        n_i^{\ov}\bff_{i}^{\ov}
        +
        \bff_{t_k,j}^{\ov}
    }{
        n_i^{\ov}+1
    },
    \label{eq:clip-running-average}
\end{equation}
after which $n_i^{\ov}$ is incremented. 

\subsection{Fusing Source Tracks into 3D}
\label{subsec:source_track_fusion}

As illustrated by the reconstruction path in~\cref{fig:trackgraph-main-figure}, source track masks are fused into the \tsdf{} and mesh. Each voxel $v$ stores at most $K$ source track hypotheses in $H_v=\{(\tau_k,\ell_{v,\tau_k})\}_{k=1}^{K}$, where $\ell_{v,\tau_k}$ is the accumulated weight of $\tau_k$ in voxel $v$. For each voxel $v$, we project its position into the current image using the estimated camera pose and intrinsics. If the projected voxel lies within the mask of source track $\tau_i$, its weight is updated as
\begin{equation}
    \ell_{v,\tau_i}
    =
    \ell_{v,\tau_i}
    +
    w_{\mathrm{TSDF}}\gamma_t(\tau_i),
    \label{eq:evidence_update}
\end{equation}
where $w_{\mathrm{TSDF}}$ is the geometric observation weight used by 
the \tsdf{} fusion. Because patch-level propagation produces coarser masks with uneven boundaries that may persist after an object leaves the image, we assign propagated observations lower confidences that decay with time since the last keyframe correction:
\begin{equation}
    \gamma_t(\tau)
    =
    \begin{cases}
        1, & \text{for a keyframe mask},\\
        \dfrac{\alpha_{\mathrm{p}}}{1+\beta_{\mathrm{p}}\Delta_t(\tau)},
        & \text{for a propagated mask},
    \end{cases}
    \label{eq:source_track_confidence}
\end{equation}
where $\alpha_{\mathrm{p}}$ scales propagation confidence, $\beta_{\mathrm{p}}$ controls its decay, and $\Delta_t(\tau)$ counts frames since the latest FastSAM keyframe detection assigned to $\tau$. 

An unrepresented source track fills an empty slot or, if all $K$ slots are occupied, replaces the hypothesis with the lowest accumulated weight. This lets new tracks enter the bounded voxel state without requiring their initial weight to outcompete existing hypotheses. Displaced hypotheses are nevertheless accumulated into 3D segments, described further in \cref{subsec:segment-construction}.

During mesh extraction via Marching Cubes, each mesh node inherits the hypothesis set $H_v$ of its supporting voxel $v$ into its attribute $H$. Winning identities determine how vertices are assigned to 3D segments, while non-winning hypotheses are retained to determine overlap between tracks.

\subsection{3D Segment Updater}
\label{subsec:segment-construction}
The segment updater, shown in the upper right of~\cref{fig:trackgraph-main-figure}, converts source tracks into persistent 3D segment nodes. Mesh vertices within the active window are grouped according to their winning source track identity $\tau_j^\star=\arg\max_{\tau_k}\ell_{j,\tau_k}$, where $\ell_{j,\tau_k}$ is the inherited weight of source track $\tau_k$ at mesh vertex $j$. When an active track $\tau_q$ is represented in the mesh but is not yet assigned to a segment, we compare it to existing active segment nodes. If a segment match is found, $\tau_q$ is assigned to that node; otherwise, a new segment node is created. This allows multiple source tracks corresponding to the same physical segment to be consolidated into a single 3D segment, for example after a tracking interruption.

When a source track $\tau_q$ has not yet been assigned to a 3D segment, we search for an existing active segment that occupies the same region of the mesh. Each mesh vertex $ N_{M}$ stores several source track hypotheses in $H$. Therefore, if $\tau_q$ is fused into vertices that also contain hypotheses of tracks already belonging to an existing segment, we treat that segment as a candidate and add it to the candidate set $\mathcal{C}_q$. For each candidate segment $\node{\segment^i}{}\in\mathcal{C}_q$, we compare the 3D mesh vertices which $\tau_q$ is fused into with those of the segment. Let $\mathcal{V}_q$ be the set of active mesh vertices whose hypothesis sets contain $\tau_q$, and let $\mathcal{V}_i$ be the vertices whose hypothesis sets contain any source track already assigned to segment $\node{\segment^i}{}$. We compute the overlap in both directions as

\begin{equation}
    \rho_{q\rightarrow i}
    =
    \frac{
        |\mathcal{V}_q\cap\mathcal{V}_i|
    }{
        |\mathcal{V}_q|
    },
    \qquad
    \rho_{i\rightarrow q}
    =
    \frac{
        |\mathcal{V}_q\cap\mathcal{V}_i|
    }{
        |\mathcal{V}_i|
    }.
    \label{eq:short-term-overlap}
\end{equation}
The ratios measure the fraction of vertices in $\mathcal{V}_q$ that also belong to $\mathcal{V}_i$ and vice versa, and a candidate is accepted if either ratio is at least $\theta_{\rho}$. If both ratios are high, this indicates near-duplicate segments, while one high and one low would indicate containment from oversegmentation. The DINOv3 feature similarity between the track and candidate node must also exceed a threshold of $\theta_{\mathrm{sim}}$.

Among candidates, we prioritize mutual overlap over containment, then rank by shared vertex count and DINOv3 similarity.
If no match is found, a new node is created for $\tau_q$. Multiple source tracks may thus be consolidated into one 3D segment after tracking interruptions. We then update the segment's associated mesh vertices, bounding box $\bbox{i}$, centroid $\point{i}{S}$, and DINOv3 feature $\bff_{i}^{\dino}$, with the source track CLIP feature $\bff_{q}^{\ov}$ added to $\bF_{i}^{\ov}$.

\subsection{Long-term Re-identification}
\label{subsec:segment-association}
The active-window association described above relies on source track weights co-occurring on the same mesh vertices. Once a segment leaves the active window, a later observation reconstructs its surface using new mesh vertices that share no source track weights with the historical geometry. The active window track-to-segment mechanism can therefore no longer associate the observations, potentially creating duplicate nodes for the same physical object. Long-term re-identification (lower part of~\cref{fig:trackgraph-main-figure}) addresses these revisits.

We maintain a k-d tree over the centroids of historical segment nodes. For an active query segment $\node{\segment^q}{}$, a centroid search returns a set $\mathcal{H}_q$ of nearby historical candidates, avoiding comparison with the complete segment layer. For each candidate $\node{\segment^h}{}\in\mathcal{H}_q$, let $S_q$ and $S_h$ denote the 3D positions of the mesh vertices connected to the query and historical segments, respectively. Following ConceptGraphs~\cite{Gu2024conceptgraphs} and HOV-SG~\cite{werby2023hovsg}, we compute the fraction of query vertices lying within $r_{\mathrm{nn}}$ of any vertex of a historical segment,
\begin{equation}
    o_{qh}
    =
    \frac{
        \left|
        \left\{
            \bx\in S_q
            \;\middle|\;
            \min_{\by\in S_h}
            \norm{\bx-\by}_2
            \leq r_{\mathrm{nn}}
        \right\}
        \right|
    }{
        |S_q|
    }.
    \label{eq:historical-overlap}
\end{equation}

We combine this geometric overlap with the cosine similarity between the segments' aggregated DINOv3 features,
\begin{equation}
    J_{qh}
    =
    \lambda_o o_{qh}
    +
    \lambda_a
    \cos\!\left(
        \feature{q}{\dino},
        \feature{h}{\dino}
    \right),
    \,
    \lambda_o+\lambda_a=1.
    \label{eq:historical-score}
\end{equation}
Among candidates whose geometric overlap, appearance similarity, and combined score pass their respective thresholds, we select the candidate with the highest $J_{qh}$.

When an association is accepted, the active node is merged into the historical node. Its source tracks and open-vocabulary features are transferred, and its mesh-to-segment edges are redirected to the historical node. The historical node thereby becomes active again with the newly reconstructed geometry. If no candidate satisfies the association criteria, the active segment remains distinct.

% !TeX root = ../../main.tex

\begin{figure*}[!t]
    \centering
    \scriptsize
    \setlength{\tabcolsep}{1.5pt}
    \begin{tabular}{@{}cccccc@{}}
        \textbf{\methodtitle} &
        \textbf{\methodtitle~(ViT-L)} &
        \textbf{ConceptGraphs} &
        \textbf{HOV-SG} &
        \textbf{FindAnything (ViT-L)} &
        \textbf{OVI-MAP (SigLIP-L)} \\
        \includegraphics[width=0.161\textwidth]{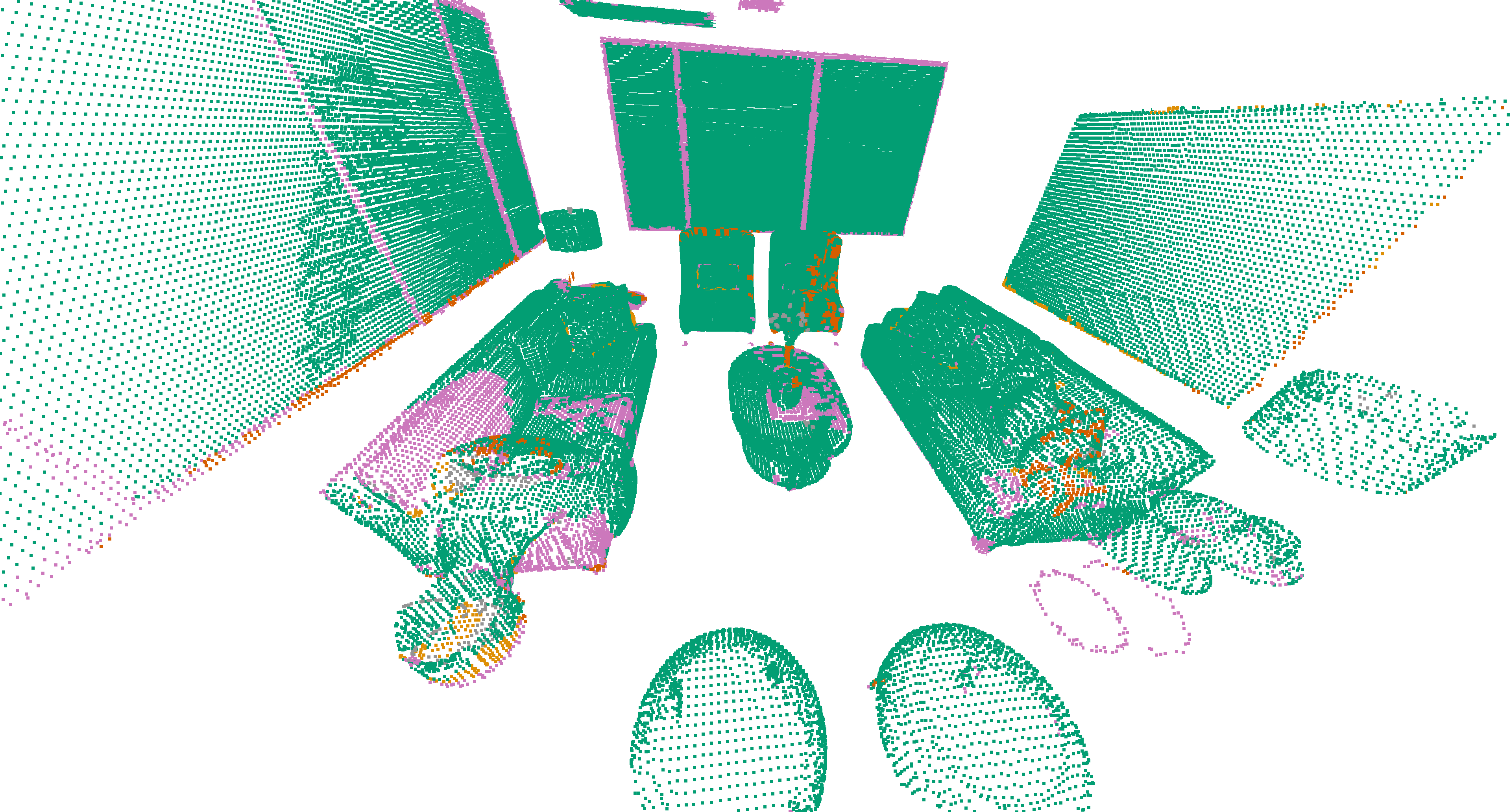} &
        \includegraphics[width=0.161\textwidth]{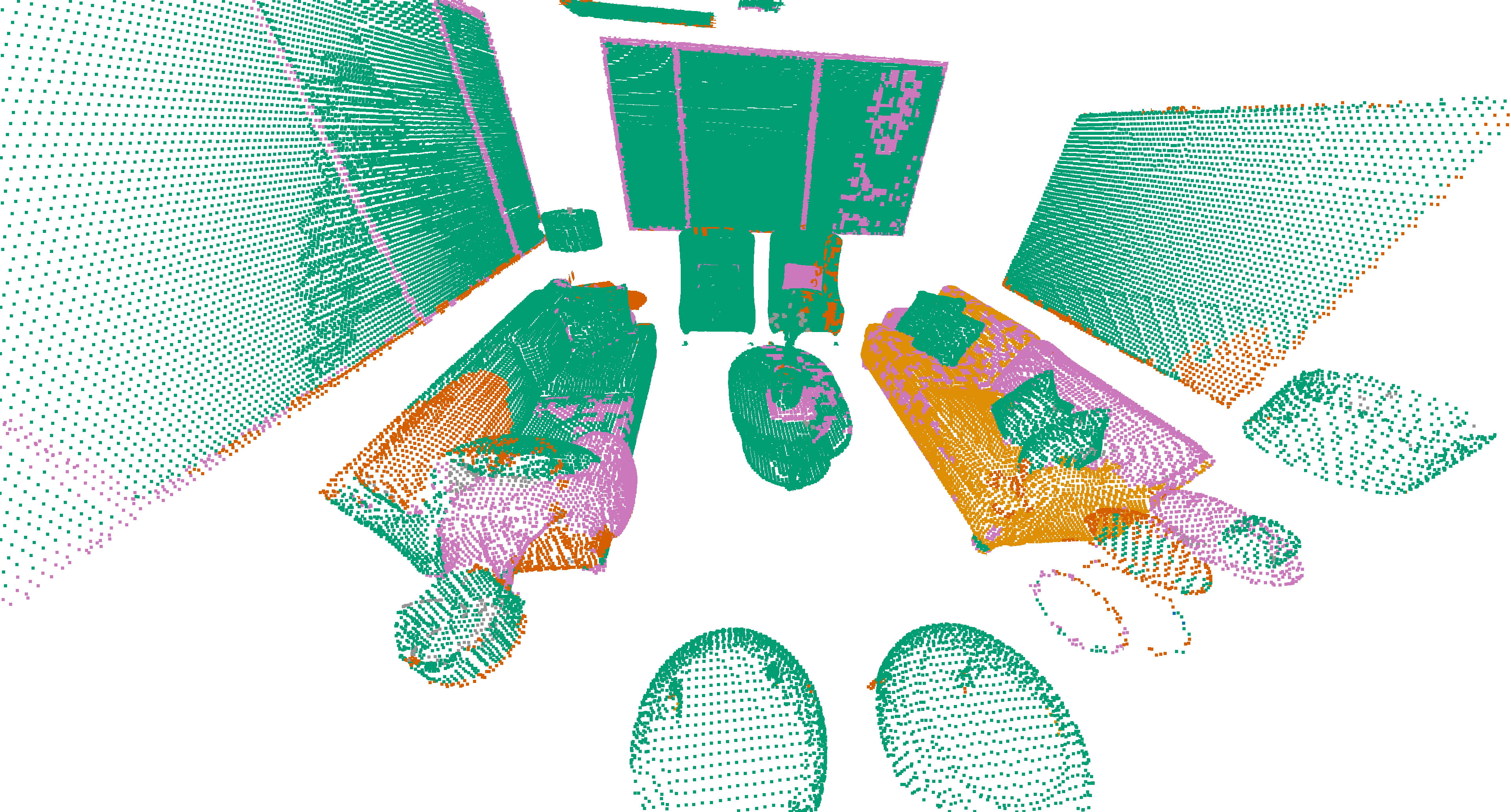} &
        \includegraphics[width=0.161\textwidth]{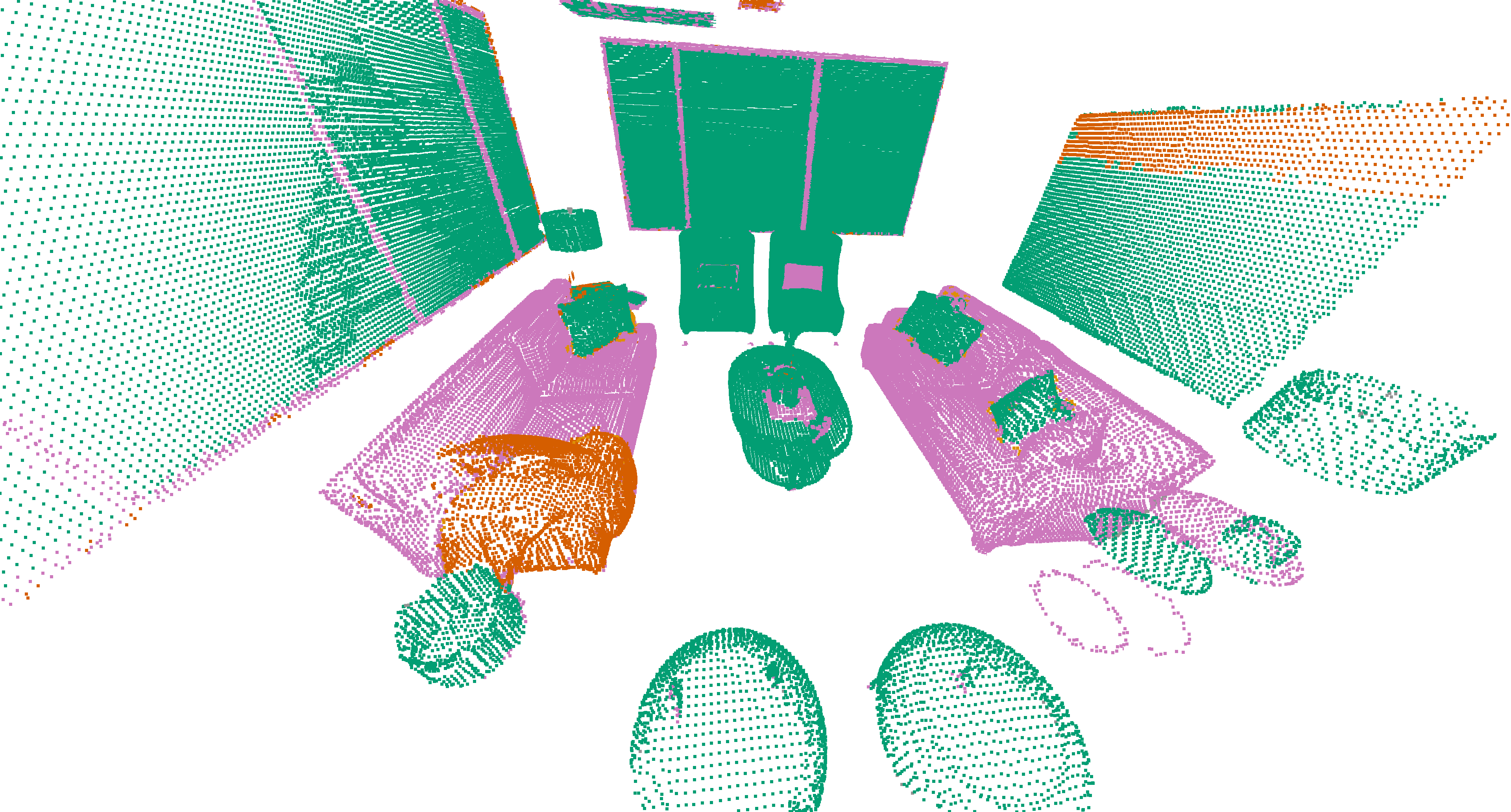} &
        \includegraphics[width=0.161\textwidth]{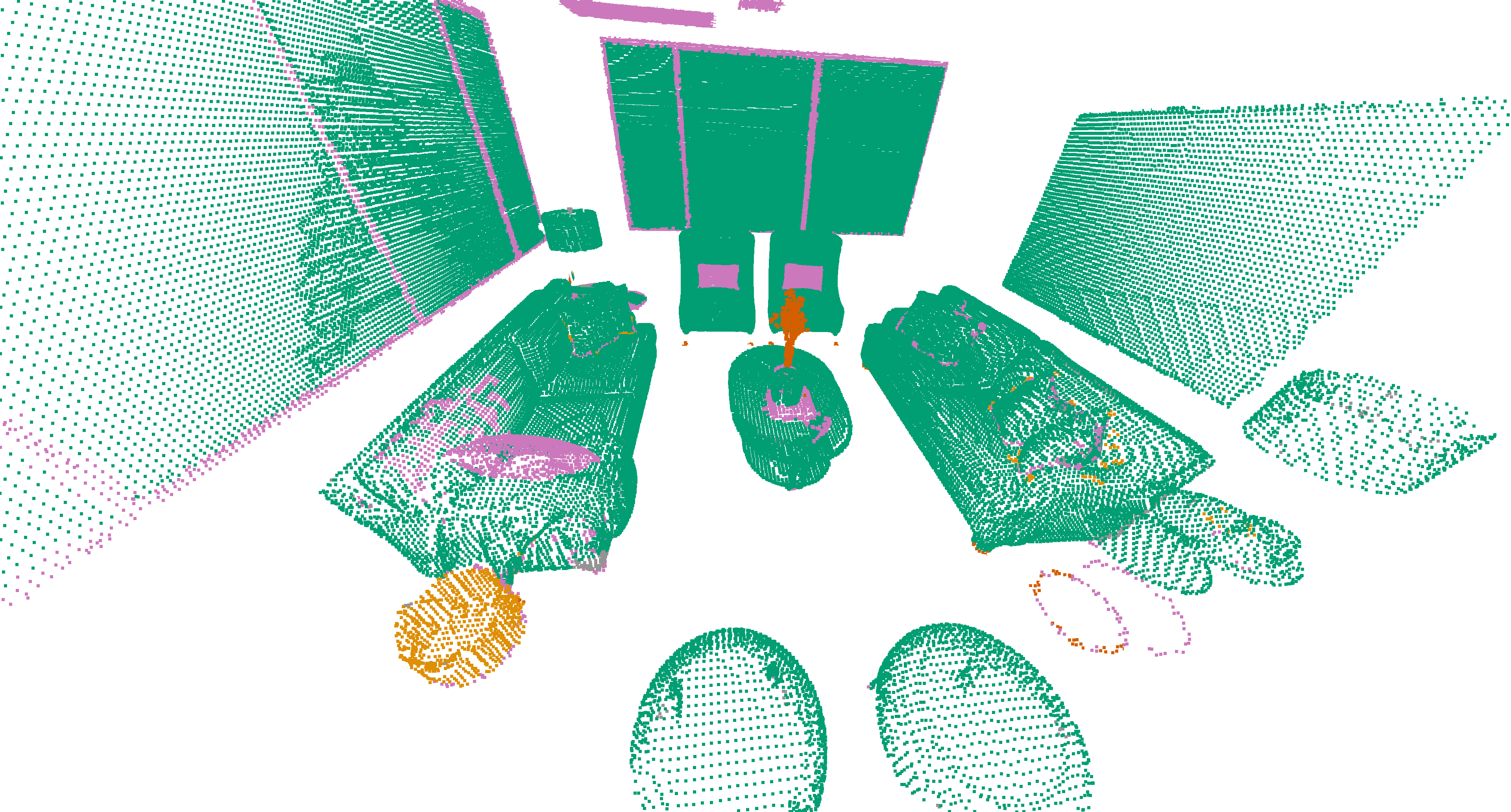} &
        \includegraphics[width=0.161\textwidth]{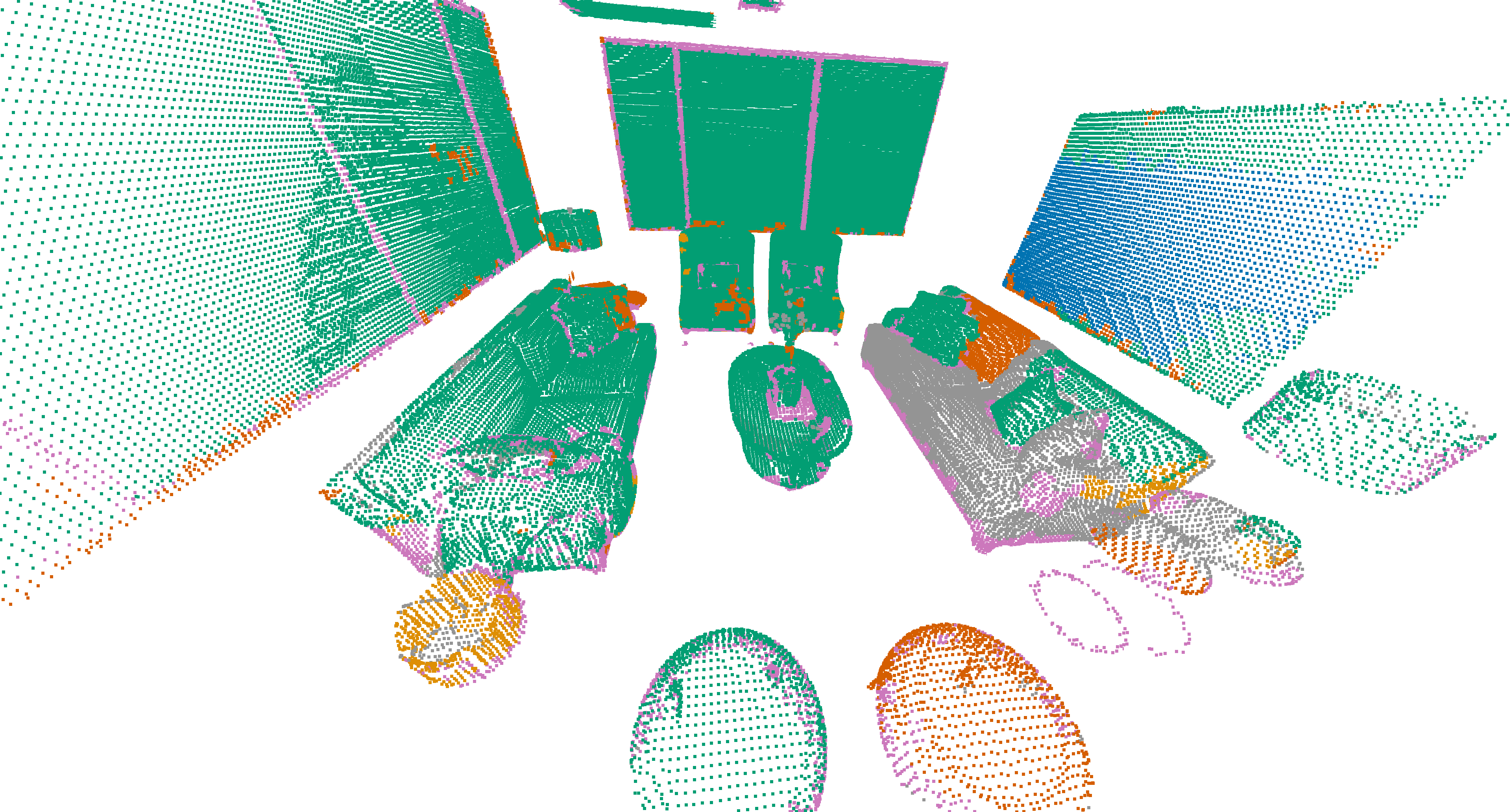} &
        \includegraphics[width=0.161\textwidth]{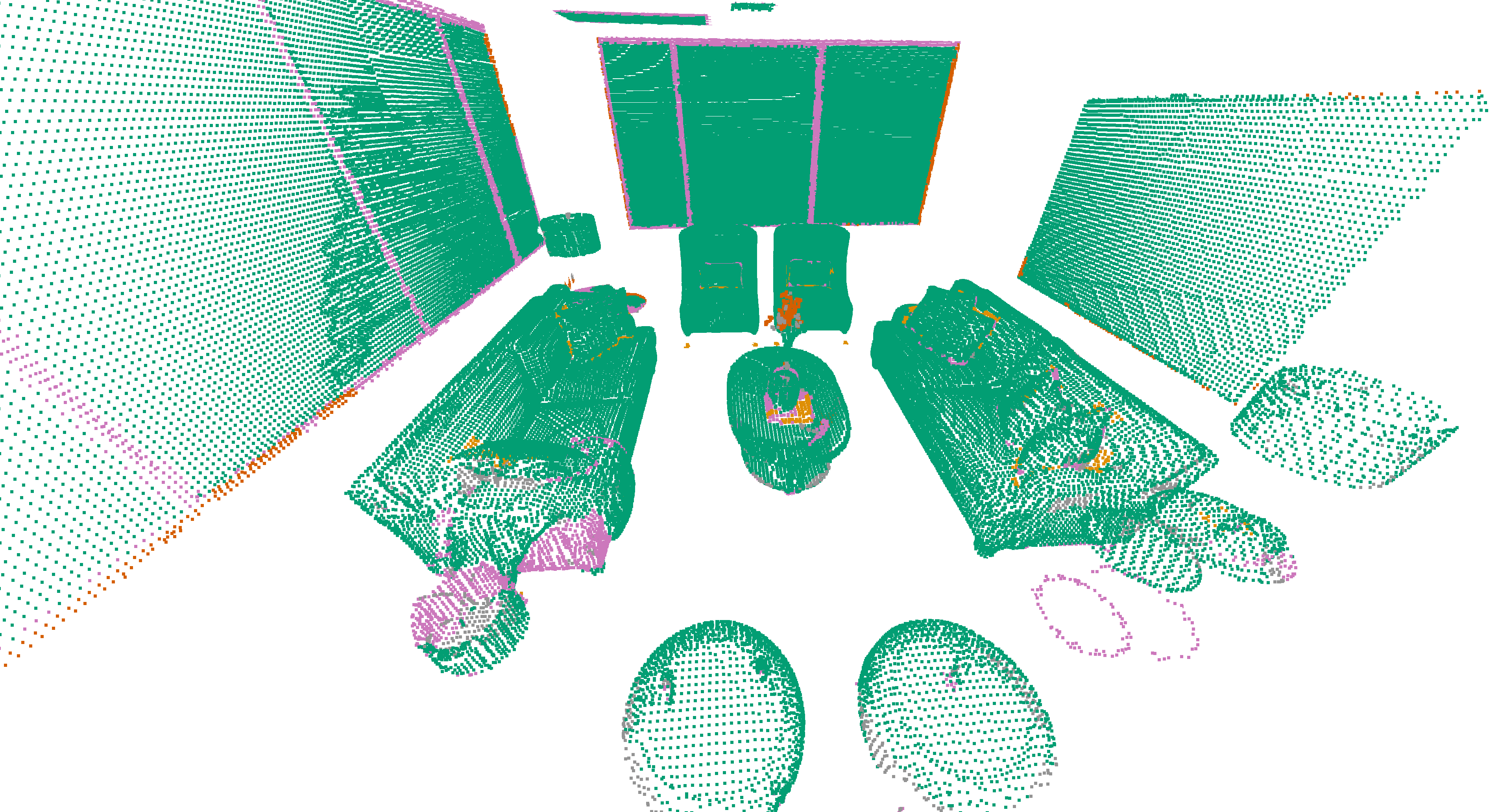} \\
    \end{tabular}

    \vspace{0.35em}
    \begingroup
    \setlength{\fboxsep}{0pt}
    \newcommand{\openlexkey}[2]{%
        \colorbox[RGB]{#1}{\phantom{\rule{1.8mm}{1.8mm}}}\,#2}
    \openlexkey{2,158,115}{synonym}\hspace{1.0em}
    \openlexkey{1,115,178}{depiction}\hspace{1.0em}
    \openlexkey{222,143,5}{visually similar}\hspace{1.0em}
    \openlexkey{204,120,188}{clutter}\hspace{1.0em}
    \openlexkey{148,148,148}{missing}\hspace{1.0em}
    \openlexkey{213,94,0}{incorrect}
    \endgroup
    \vspace{-1em}
    \caption{OpenLex3D Top-5 visualizations for Replica room0. White points are ignored, background, or non-evaluated areas. \methodtitle{} maintains broad synonym coverage comparable to HOV-SG \cite{werby2023hovsg} and OVI-MAP \cite{deng_ovi-mapopen-vocabulary_2026}, while ConceptGraphs \cite{Gu2024conceptgraphs} and FindAnything \cite{laina_findanything_2026} exhibit larger contiguous clutter, missing, or incorrect regions. }
    \label{fig:openlex3d-replica-category-visualizations}
    \vspace{-2em}
\end{figure*}

\subsection{Open-Vocabulary Galleries and Querying}
\label{subsec:ov-querying}

Each segment node stores an open-vocabulary feature gallery $\featurematrix{i}{\ov}=\{(\tau,\feature{\tau}{\ov})\mid\tau\in T_i\}$ keyed by its assigned source tracks, where $\feature{\tau}{\ov}$ is the aggregated CLIP feature of source track $\tau$. When a source track is assigned to the node, its current feature is added to the gallery, while subsequent feature updates keep the corresponding gallery entry synchronized with the track average from~\cref{eq:clip-running-average}.

Features from different source tracks are retained separately rather than averaged into a single node-level feature. Different tracks may capture distinct views of the same physical object, and averaging CLIP features across views can reduce classification accuracy~\cite{kassab2024barenecessities}. The gallery thus preserves these alternatives while storing only one feature per source track rather than one per keyframe.

\mypar{Querying the Graph} A text or image query is embedded into the same CLIP space as the stored features. Given a normalized query feature $\bq$, the score of segment node $\node{i}^{\segment}$ is the maximum cosine similarity across its gallery,
\begin{equation}
    s_i(\bq)
    =
    \max_{
        (\tau,\feature{\tau}{\ov})
        \in
        \featurematrix{i}{\ov}
    }
    \dfrac{\bq \cdot \feature{\tau}{\ov}}{\left\| \bq \right\| \left\|\feature{\tau}{\ov} \right\|}.
    \label{eq:query_score}
\end{equation}
The maximum selects the view that best matches the query, avoiding dilution from less informative views. The query interface returns a ranked list of segment nodes.

% !TeX root = ../main.tex

\section{EXPERIMENTAL RESULTS}\label{sec:results}

We evaluate open-vocabulary segmentation and object retrieval on Replica~\cite{replica19arxiv}, ScanNet++~\cite{yeshwanth2023scannet++}, and HM3D~\cite{Yadav2023HM3DSem} following OpenLex3D~\cite{kassab2025openlex3d}. Runtime and resource usage are evaluated on Replica, whose scenes have comparable durations and memory footprints and for which baseline measurements are available \cite{martins2024ovoslam}. We compare our framework against strong baselines such as ConceptGraphs \cite{Gu2024conceptgraphs}, HOV-SG~\cite{werby2023hovsg}, OVI-MAP~\cite{deng_ovi-mapopen-vocabulary_2026}, and FindAnything~\cite{laina_findanything_2026}. Real-world experiments demonstrate onboard quadruped deployment and mapping and retrieval from recorded drone data.

\subsection{Experimental Setup} 
\label{sec:experiments_setup}
\mypar{Implementation Details} For our main results, \method{} computes FastSAM-x masks and OpenCLIP ViT-H/14~\cite{schuhmann_laion-400m_2021} features every $K_f=24$ frames. We choose the DINOv3 ViT-S+ \cite{simeoni_dinov3_2025} variant due to its high-quality visual embeddings packaged in a small 29M-parameter model, ideal for online propagation. Input images to DINOv3 are resized to 720 pixels along the short side, while retaining the original aspect ratio. For keyframe association (see \cref{eq:keyframe_assoc}), we set $w_{\mathrm{IoU}}=w_{\mathrm{feat}}=0.5$, require $\mathrm{IoU}\geq0.15$ and $s_{ij}\geq0.35$, and use $\eta=0.2$. During \tsdf~fusion, we set $\alpha_{\mathrm{p}}=0.25$ and $\beta_{\mathrm{p}}=0.1$ in \cref{eq:source_track_confidence} to downweight propagated observations and decay their influence between keyframes. Each voxel of size $0.03\mathrm{m}$ maintains up to $K=4$ source track hypotheses. The 3D segment updater creates a node when a source track has at least $20$ vertices, and for the active-window overlap criterion of \cref{eq:short-term-overlap}, we require $|\mathcal{V}_q\cap\mathcal{V}_i|\geq5$, $\theta_{\rho}=0.75$, and $\theta_{\mathrm{sim}}=0.80$. For long-term re-identification, we set $r_{\mathrm{nn}}=0.10\mathrm{m}$ and $\lambda_o=\lambda_a=0.5$, and require geometric overlap $o_{qh} \geq 0.42$, appearance similarity $\cos\!\left(\feature{q}{\dino},\feature{h}{\dino}\right) \geq 0.70$, and combined score $J_{qh} \geq 0.65$ (see \cref{eq:historical-overlap,eq:historical-score}). Hyperparameters were selected empirically through preliminary experiments and kept fixed across all benchmark datasets.

% !TeX root = ../../main.tex

\begin{table}[!t]
    \centering
    \scriptsize
    \renewcommand{\arraystretch}{1.12}
    \setlength{\tabcolsep}{2.0pt}
    \begin{adjustbox}{max width=1\columnwidth}
    \begin{tabular}{ll
        >{\columncolor{openlexS}}c
        c
        c
        c
        c
        >{\columncolor{openlexI}}c}
        \toprule
        \textbf{Data} & \textbf{Method}
        & \seghead{openlexSHead}{S $\uparrow$}
        & \seghead{openlexDHead}{D}
        & \seghead{openlexVSHead}{VS}
        & \seghead{openlexCHead}{C $\downarrow$}
        & \seghead{openlexMHead}{M $\downarrow$}
        & \seghead{openlexIHead}{I $\downarrow$} \\
        \midrule
        Replica & ConceptGraphs$^{\dagger}$ & 0.41 & 0.01 & 0.11 & 0.24 & 0.02 & 0.22 \\
        & HOV-SG$^{\dagger}$ & 0.45 & 0.00 & 0.05 & 0.27 & 0.07 & 0.16 \\
        & FindAnything (ViT-L) & 0.41 & 0.03 & 0.06 & 0.23 & 0.08 & 0.19 \\
        & OpenMask3D & 0.43 & 0.01 & 0.07 & 0.29 & 0.10 & \textbf{0.10} \\
        & ConceptFusion & 0.32 & 0.01 & 0.09 & \textbf{0.16} & \textbf{0.00} & 0.41 \\
        & OpenScene (ViT-L OpenSeg) & 0.44 & 0.00 & 0.06 & 0.30 & 0.07 & 0.13 \\
        & OVI-MAP (SigLIP-L)$^{\dagger}$ & 0.49 & 0.00 & 0.07 & 0.26 & 0.05 & 0.12 \\
        & OVI-MAP$^{\dagger}$ & 0.40 & 0.01 & 0.10 & 0.28 & 0.05 & 0.16 \\
        \cmidrule(lr){2-8}
        \cellcolor{openlexRow}
        & \cellcolor{openlexRow}\textbf{\methodtitle~(Ours)}
        & \textbf{0.50} & 0.00 & 0.07 & 0.23 & 0.02 & 0.17 \\
        \cellcolor{openlexRow}
        & \cellcolor{openlexRow}\textbf{\methodtitle~(Ours, ViT-L)}
        & 0.47 & 0.01 & 0.05 & 0.23 & 0.02 & 0.22 \\
        \midrule
        ScanNet++ & ConceptGraphs$^{\dagger}$ & 0.26 & 0.02 & 0.05 & 0.10 & 0.13 & 0.44 \\
        & HOV-SG$^{\dagger}$ & \textbf{0.40} & 0.02 & 0.04 & 0.16 & 0.08 & 0.30 \\
        & OpenMask3D & 0.27 & 0.01 & 0.03 & 0.29 & 0.13 & \textbf{0.27} \\
        & ConceptFusion & 0.29 & 0.01 & 0.03 & \textbf{0.08} & \textbf{0.04} & 0.54 \\
        & OpenScene (ViT-L OpenSeg) & 0.16 & 0.00 & 0.02 & 0.23 & 0.22 & 0.36 \\
        & OVI-MAP (SigLIP-L)$^{\dagger}$ & 0.29 & 0.01 & 0.03 & 0.23 & 0.14 & 0.29 \\
        & OVI-MAP$^{\dagger}$ & 0.25 & 0.01 & 0.03 & 0.22 & 0.14 & 0.36 \\
        \cmidrule(lr){2-8}
        \cellcolor{openlexRow}
        & \cellcolor{openlexRow}\textbf{\methodtitle~(Ours)}
        & 0.35 & 0.01 & 0.03 & 0.15 & 0.17 & 0.29 \\
        \cellcolor{openlexRow}
        & \cellcolor{openlexRow}\textbf{\methodtitle~(Ours, ViT-L)}
        & 0.31 & 0.01 & 0.03 & 0.14 & 0.17 & 0.33 \\
        \midrule
        HM3D & ConceptGraphs$^{\dagger}$ & 0.27 & 0.02 & 0.03 & 0.12 & 0.08 & 0.47 \\
        & HOV-SG$^{\dagger}$ & 0.33 & 0.02 & 0.04 & 0.18 & 0.08 & 0.36 \\
        & FindAnything (ViT-L) & 0.33 & 0.02 & 0.02 & 0.12 & 0.14 & 0.37 \\
        & OpenMask3D & 0.31 & 0.01 & 0.03 & 0.13 & 0.26 & \textbf{0.26} \\
        & ConceptFusion & 0.23 & 0.01 & 0.03 & \textbf{0.09} & 0.08 & 0.57 \\
        & OpenScene (ViT-L OpenSeg) & 0.18 & 0.00 & 0.02 & 0.16 & 0.06 & 0.59 \\
        & OVI-MAP (SigLIP-L)$^{\dagger}$ & \textbf{0.37} & 0.01 & 0.04 & 0.17 & 0.08 & 0.33 \\
        & OVI-MAP$^{\dagger}$ & 0.32 & 0.01 & 0.03 & 0.21 & 0.08 & 0.34 \\
        \cmidrule(lr){2-8}
        \cellcolor{openlexRow}
        & \cellcolor{openlexRow}\textbf{\methodtitle~(Ours)}
        & 0.32 & 0.01 & 0.03 & 0.22 & 0.04 & 0.39 \\
        \cellcolor{openlexRow}
        & \cellcolor{openlexRow}\textbf{\methodtitle~(Ours, ViT-L)}
        & 0.24 & 0.01 & 0.03 & 0.22 & \textbf{0.03} & 0.47 \\
        \bottomrule
    \end{tabular}
    \end{adjustbox}

    \vspace{0.2em}
    \parbox{\columnwidth}{\scriptsize
    \, S: synonyms, D: depictions, VS: visually similar, C: clutter, M: missing, I: incorrect}
    \caption{OpenLex3D Top-5 segmentation unit-scale frequencies. \vl~encoders other than ViT-H are named in parentheses, and $^{\dagger}$ indicates that only every 10th frame is processed. \methodtitle~achieves competitive results compared to baselines across datasets, demonstrating the best and second-best synonyms score on Replica \cite{replica19arxiv} and ScanNet++~\cite{yeshwanth2023scannet++}, respectively.}
    \label{tab:openlex3d-system-segmentation}
    \vspace{-2em}
\end{table}

% !TeX root = ../../main.tex

\begin{table}[!t]
    \centering

    \scriptsize
    \renewcommand{\arraystretch}{1.12}
    \setlength{\tabcolsep}{5.0pt}
    \begin{adjustbox}{max width=0.98\columnwidth}
    \begin{tabular}{llccc}
        \toprule
        \textbf{Data} & \textbf{Method}
        & \textbf{AP $\uparrow$} & \textbf{AP50 $\uparrow$}
        & \textbf{AP25 $\uparrow$} \\
        \midrule
        Replica & ConceptGraphs$^{\dagger}$ & 5.86 & 11.32 & 22.39 \\
        & HOV-SG$^{\dagger}$ & 5.76 & 11.67 & 25.30 \\
        & FindAnything (ViT-L) & 2.28 & 6.00 & 13.87 \\
        & OVI-MAP (SigLIP-L)$^{\dagger}$ & \textbf{10.49} & \textbf{19.40} & \textbf{35.16} \\
        & OVI-MAP$^{\dagger}$ & 7.83 & 14.21 & 26.79 \\
        \cmidrule(lr){2-5}
        \rowcolor{openlexRow}
        & \textbf{\methodtitle~(Ours)} & 6.00 & 15.78 & 30.81 \\
        \rowcolor{openlexRow}
        & \textbf{\methodtitle~(Ours, ViT-L)} & 3.94 & 9.90 & 23.84 \\
        \midrule
        ScanNet++ & ConceptGraphs$^{\dagger}$ & 1.45 & 4.36 & 15.27 \\
        & HOV-SG$^{\dagger}$ & 1.79 & 4.95 & \textbf{18.75} \\
        & OVI-MAP (SigLIP-L)$^{\dagger}$ & \textbf{3.59} & \textbf{7.78} & 17.75 \\
        & OVI-MAP$^{\dagger}$ & 3.20 & 6.94 & 14.82 \\
        \cmidrule(lr){2-5}
        \rowcolor{openlexRow}
        & \textbf{\methodtitle~(Ours)} & 2.21 & 5.55 & 17.15 \\
        \rowcolor{openlexRow}
        & \textbf{\methodtitle~(Ours, ViT-L)} & 2.21 & 5.41 & 15.03 \\
        \midrule
        HM3D & ConceptGraphs$^{\dagger}$ & 5.09 & 8.05 & 11.18 \\
        & HOV-SG$^{\dagger}$ & 3.44 & 5.39 & 7.42 \\
        & FindAnything (ViT-L) & 1.56 & 3.60 & 6.81 \\
        & OVI-MAP (SigLIP-L)$^{\dagger}$ & \textbf{8.96} & \textbf{13.67} & \textbf{18.07} \\
        & OVI-MAP$^{\dagger}$ & 6.89 & 10.59 & 13.46 \\
        \cmidrule(lr){2-5}
        \rowcolor{openlexRow}
        & \textbf{\methodtitle~(Ours)} & 2.16 & 4.55 & 10.55 \\
        \rowcolor{openlexRow}
        & \textbf{\methodtitle~(Ours, ViT-L)} & 1.47 & 3.13 & 7.36 \\
        \bottomrule
    \end{tabular}
    \end{adjustbox}

    \caption{OpenLex3D object-retrieval results in percentages. \vl~encoders other than ViT-H are named in parentheses. $^{\dagger}$ indicates that only every 10th frame is processed.  On Replica~\cite{replica19arxiv} and ScanNet++~\cite{yeshwanth2023scannet++}, \methodtitle~trails only OVI-MAP~\cite{deng_ovi-mapopen-vocabulary_2026} variants on every metric except AP25 on ScanNet++, where HOV-SG \cite{werby2023hovsg} is strongest. We trail encoder-matched baselines on HM3D AP and AP50, but are competitive on AP25.}
    \label{tab:openlex3d-system-retrieval}
    \vspace{-2.5em}
\end{table}

\mypar{Baselines Setup}
ConceptGraphs and HOV-SG segmentation and retrieval results are reported from OpenLex3D, while their efficiency values are taken from OVO-SLAM~\cite{martins2024ovoslam}, which uses an NVIDIA RTX 3090 for evaluation. \method{}, OVI-MAP, and FindAnything are evaluated on an NVIDIA A100. We evaluate FindAnything on Replica and HM3D using its standard ViT-L/14@336 configuration and generate stereo inputs following~\cite{laina_findanything_2026}; this is not possible for ScanNet++ since it is based on real-world recordings. OVI-MAP is evaluated using both its official SigLIP-L/16@384~\cite{Zhai2023Siglip} configuration, in addition to a ViT-H encoder-matched comparison. FindAnything and \method{} consume every frame, whereas ConceptGraphs, HOV-SG, and OVI-MAP process every tenth frame, reflecting standard configurations of each method~\cite{kassab2025openlex3d, deng_ovi-mapopen-vocabulary_2026}.

\mypar{Datasets and Evaluation Protocol}
The datasets span compact synthetic scenes (Replica), high-fidelity real-world reconstructions (ScanNet++), and larger, more diverse indoor environments (HM3D). For each dataset, we evaluate our method and baselines in the scenes used by OpenLex3D~\cite{kassab2025openlex3d}. OpenLex3D evaluates open-vocabulary representations without reducing their predictions to a small set of labels, providing a higher fidelity evaluation than previous methods~\cite{Gu2024conceptgraphs,werby2023hovsg, deng_ovi-mapopen-vocabulary_2026, laina_findanything_2026}. 

For semantic segmentation, we compare each point's VL feature with the text embeddings of the dataset vocabulary and select the five labels with highest cosine similarity. These labels are then matched against the ground-truth annotations to assign the point to one of the six categories in \cref{tab:openlex3d-system-segmentation}. Category frequencies are computed per ground-truth object and averaged across objects, weighting objects equally regardless of size. Floors, ceilings, and walls are excluded, and output point clouds are downsampled to $0.05$m.

For retrieval, text queries are encoded using each method’s \vl~encoder, and predictions are ranked by visual–text similarity (\method{} uses~\cref{eq:query_score}). We report \ap~averaged over 3D IoU thresholds from $0.50$ to $0.95$ in steps of $0.05$, together with AP50 and AP25 at thresholds of $0.50$ and $0.25$, respectively.

% !TeX root = ../../main.tex

\begin{table}[!t]
    \centering
    \scriptsize
    \renewcommand{\arraystretch}{1.12}
    \setlength{\tabcolsep}{1.8pt}
    \begin{tabular}{clrrrrr}
        \toprule
        & \textbf{Method} & \textbf{AP $\uparrow$} & \textbf{S $\uparrow$}
        & \textbf{Time $\downarrow$} & \textbf{RAM $\downarrow$}
        & \textbf{VRAM $\downarrow$} \\
        \midrule
        \raisebox{-25.5pt}[0pt][0pt]{\rotatebox[origin=c]{90}{ViT-H}}
        & ConceptGraphs$^{\dagger}$ & 5.86 & 0.41
        & $\sim$16\,min & 16 & 11 \\
        & HOV-SG$^{\dagger}$ & 5.76 & 0.45
        & $\sim$11\,h & 139 & 12 \\
        & OVI-MAP @ 3\,cm$^{\dagger}$ & 7.83 & 0.40 & 15.1\,min & 10.3 & 27.2 \\
        & OVI-MAP @ 5\,cm$^{\dagger}$ & 5.28 & 0.41 & 16.0\,min & 8.6 & 27.2 \\
        & \textbf{\methodtitle @ 3\,cm} & 6.00 & \textbf{0.50} & 8.9\,min & 9.5 & 8.3 \\
        & \textbf{\methodtitle @ 5\,cm} & 4.31 & 0.49 & 3.9\,min & 6.0 & 8.3 \\
        \midrule
        \raisebox{-14.5pt}[0pt][0pt]{\rotatebox[origin=c]{90}{ViT-L}}
        & FindAnything @ 3\,cm & 2.28 & 0.41 & 7.8\,min & 9.2 & 9.3 \\
        & FindAnything @ 5\,cm & 1.34 & 0.38 & 5.3\,min & \textbf{3.7} & 9.3 \\
        & \textbf{\methodtitle @ 3\,cm} & 3.94 & 0.47 & 8.4\,min & 10.3 & \textbf{7.3} \\
        & \textbf{\methodtitle @ 5\,cm} & 2.78 & 0.46 & \textbf{3.7\,min} & 5.9 & \textbf{7.3} \\
        \midrule
        & OVI-MAP (SigLIP-L) @ 3\,cm$^{\dagger}$
        & \textbf{10.49} & 0.49 & 15.0\,min & 6.7 & 27.2 \\
        & OVI-MAP (SigLIP-L) @ 5\,cm$^{\dagger}$
        & 6.93 & 0.47 & 15.7\,min & 10.9 & 27.2 \\
        \bottomrule
    \end{tabular}

    \caption{Efficiency and OpenLex3D quality on Replica. HOV-SG and ConceptGraphs efficiency values are taken from~\cite{martins2024ovoslam}, which used an NVIDIA RTX 3090. $^{\dagger}$ indicates that only every 10th frame is processed. Timings, peak RAM and VRAM in GB are averaged over all sequences. AP is in percentage points and S is the unit-scale synonym frequency.}
    \label{tab:runtime-system-comparison}
    \vspace{-2em}
\end{table}

\subsection{Open-Vocabulary Semantic Segmentation and Retrieval}
\label{sec:experiments_segmentation_retrieval}
\mypar{Semantic Segmentation}
\cref{tab:openlex3d-system-segmentation} reports the OpenLex3D Top-5 outcome frequencies, while~\cref{fig:openlex3d-replica-category-visualizations} visualizes the results for Replica's \emph{room0} scene. Across encoders, \method{} ViT-H achieves the highest synonym frequency on Replica, is second to HOV-SG on ScanNet++, and remains competitive on HM3D, where OVI-MAP with SigLIP-L performs best. Using the same \vl~encoder, \method{} achieves equal synonym scores as OVI-MAP's ViT-H variant on HM3D.  Overall, \method{} remains competitive with state-of-the-art baselines and achieves the best performance on some segmentation metrics, while using $2.4 \times$ fewer segmentation and open-vocabulary inference steps.

\mypar{Open-Vocabulary Object Retrieval}
OVI-MAP with SigLIP-L leads across all retrieval metrics and datasets except AP25 on ScanNet++, where HOV-SG using ViT-H performs best (\cref{tab:openlex3d-system-retrieval}). Among the ViT-H variants, \method{} leads in AP50 and AP25 on Replica but trails OVI-MAP in AP, and outperforms ConceptGraphs and HOV-SG in AP and AP50 on ScanNet++. With ViT-L, \method{} outperforms FindAnything on all three retrieval metrics on Replica, but only AP25 on HM3D. On HM3D, \method{} trails all baseline methods with matched encoder on AP and AP50 despite competitive point-wise semantics (\cref{tab:openlex3d-system-segmentation}) and AP25. This suggests that \method{} often retrieves the correct object region, but its segments may not fully cover the object extent in HM3D, likely due to the increased difficulty of maintaining instance continuity in larger, diverse scenes.

\begin{figure*}[!t]
    \centering
    \includegraphics[
        width=\linewidth,
        trim=0mm 0mm 0mm 2mm,
        clip
    ]{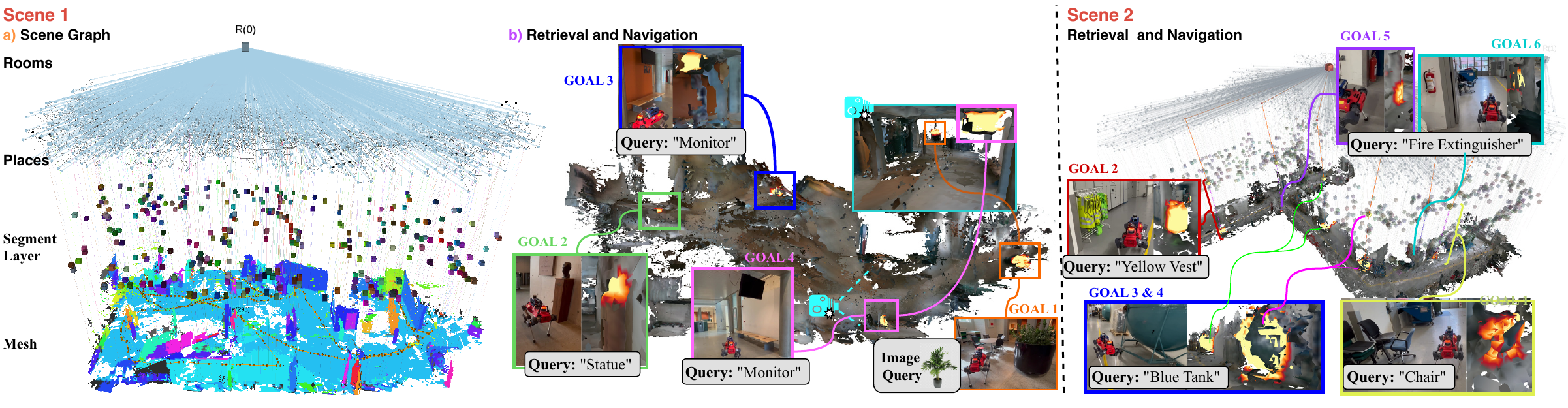}
    \caption{For two large real-world environments, \textbf{a)} shows the open-vocabulary scene graph constructed online by \methodtitle~onboard a quadruped robot, while \textbf{b)} highlights retrieval with both image- and text-based queries, with successful navigation to all segments in the indicated goal order.}
    \label{fig:real-world-experiments}
    \vspace{-4.5ex}
\end{figure*}

\subsection{Mapping Quality and Computational Efficiency}
\label{sec:experiments_runtime}
Table~\ref{tab:runtime-system-comparison} compares the most significant accuracy metrics with end-to-end sequence processing time and memory usage on Replica. Under the reported settings, \method{} has lower reported sequence processing time than ConceptGraphs, HOV-SG, and OVI-MAP, together with the lowest reported GPU memory usage. As discussed in \cref{sec:experiments_setup}, these efficiency results are not all measured under identical hardware and input-frame settings and should therefore be interpreted as indicative comparisons. That said, the bulk of HOV-SG processing is on the CPU, while the GPU is reserved for 2D segmentation inference. Meanwhile, \method~CPU memory depends strongly on voxel resolution: increasing the voxel size from $0.03$\,m to $0.05$\,m reduces peak RAM usage by more than one-third, with little change in synonym frequency but lower retrieval AP. 

Specifically, FindAnything~\cite{laina_findanything_2026}, the closest literature result in our domain, associates per-frame segments with rendered occupancy submaps, whereas \method{} propagates identities in image space and builds a hierarchical scene graph. Our evaluation covers more datasets and a broader set of segmentation, retrieval, and efficiency metrics. In terms of robot evaluation, FindAnything integrates semantic exploration on a drone, whereas we deploy \method{} onboard a quadruped and additionally evaluate mapping and retrieval on recorded aerial data. With ViT-L, \method{} performs better on Replica retrieval but worse on HM3D AP and AP50, while being faster on Replica with lower GPU but higher CPU memory at $5\,\mathrm{cm}$.
\subsection{Ablations}
\mypar{Propagation-based Tracking} To evaluate propagation between sparse keyframes, we replace our tracker with per-frame FastSAM detections associated by BoT-SORT~\cite{Aharon2022BoTSORTRA}, keeping the remaining \method~pipeline fixed. \cref{tab:tracking-ablation-quality-cost} shows that our tracker sustains $15.0$\,Hz versus $6.0$\,Hz ($2.5\times$) on Replica, with nearly identical AP and higher AP50, AP25, and synonym frequency. On Replica, this indicates that DINOv3 propagated masks preserve mapping quality while increasing throughput.

% !TeX root = ../../main.tex

\begin{table}[h!]
    \vspace{-0.5em}
    \centering
    \scriptsize
    \renewcommand{\arraystretch}{1.12}
    \setlength{\tabcolsep}{1.8pt}
    \begin{adjustbox}{max width=0.98\columnwidth}
    \begin{tabular}{lrrrrrrr}
        \toprule
        \textbf{Tracker} & \textbf{AP $\uparrow$} & \textbf{AP50 $\uparrow$}
        & \textbf{AP25 $\uparrow$} & \textbf{S $\uparrow$}
        & \textbf{Max input Hz $\uparrow$} & \textbf{RAM $\downarrow$}
        & \textbf{VRAM $\downarrow$} \\
        \midrule
        FastSAM + BoT-SORT & \textbf{6.01} & 13.74 & 29.46 & 0.48 & 6.0 & 13.2
        & \textbf{7.1} \\
        \rowcolor{openlexRow}
        \textbf{DINOv3-based (Ours)} & 6.00 & \textbf{15.78} & \textbf{30.81} & \textbf{0.50}
        & \textbf{15.0} & \textbf{9.5} & 8.3 \\
        \bottomrule
    \end{tabular}
    \end{adjustbox}

    \vspace{0.3em}
      \caption{Tracker ablation on Replica (ViT-H, 3\,cm voxels). Max input Hz is the highest rate sustained without dropping frames.}
    \label{tab:tracking-ablation-quality-cost}
    \vspace{-3em}
\end{table}

\mypar{CLIP Feature Galleries} We evaluate the gallery design in \cref{subsec:ov-querying} by replacing max-gallery scoring with cosine similarity to each segment's normalized mean CLIP feature. The results of \cref{tab:gallery-ablation} support retaining feature galleries.
\begin{table}[h!]
    \centering
    \scriptsize
    \renewcommand{\arraystretch}{1.0}
    \setlength{\tabcolsep}{3.2pt}
    \begin{adjustbox}{max width=0.98\columnwidth}
    \begin{tabular}{lrrrr}
        \toprule
        \textbf{Scoring}
        & \textbf{AP $\uparrow$}
        & \textbf{AP50 $\uparrow$}
        & \textbf{AP25 $\uparrow$}
        & \textbf{S $\uparrow$} \\
        \midrule
        Mean feature
        & 5.65 & 15.09 & 29.41 & 0.45 \\
        \rowcolor{openlexRow}
        \textbf{Max gallery}
        & \textbf{6.00} & \textbf{15.78}
        & \textbf{30.81} & \textbf{0.50} \\
        \bottomrule
    \end{tabular}
    \end{adjustbox}
    \caption{Gallery ablation on Replica with ViT-H, averaged over all scenes. AP is in percentage points; S is the unit-scale synonym frequency.}
    \label{tab:gallery-ablation}
    \vspace{-9ex}
\end{table}
\subsection{Real-world Experiments}
\label{sec:experiments_real_world}
We deploy \method{} on a quadruped robot equipped with an Intel RealSense D455 RGB-D camera, an Ouster OS0 LiDAR, and a VectorNav VN-100 IMU. The LiDAR and IMU provide odometry~\cite{Dharmadhikari2026uas}, and all \method{} components run onboard an NVIDIA Jetson AGX Thor. Additionally, we test \method{} on recorded RGB-D data captured by a drone in multiple indoor environments.

\mypar{Configuration} 
We retain FastSAM-x and DINOv3 ViT-S+ from~\cref{sec:experiments_setup}, but use the lighter CLIP ViT-B encoder and resize DINOv3 inputs to 480 pixels along the short side for efficient onboard processing. We use $0.125$\,m TSDF voxels and process every fourth frame of the $30$\,Hz camera stream, yielding a nominal $7.5$\,Hz rate for both instance tracking and scene graph construction.

\mypar{Quadruped Autonomous Exploration and Object Search} 
We evaluate the complete system in three large-scale indoor environments (\cref{fig:intro,fig:real-world-experiments}). Each experiment comprises two autonomous phases: exploration using a volumetric planner~\cite{zacharia2026omniplanner} while constructing the open-vocabulary scene graph, followed by object search using text or reference-image queries. Using retrieved segment positions as navigation goals, the quadruped robot searches for individual objects and multiple instances of the same category. Across all three environments, all queried targets were successfully retrieved.

\begin{figure}[t]
    \vspace{0.5em}
    \centering
    \includegraphics[
        width=\columnwidth,
        trim=0mm 0mm 10mm 0mm,
        clip
    ]{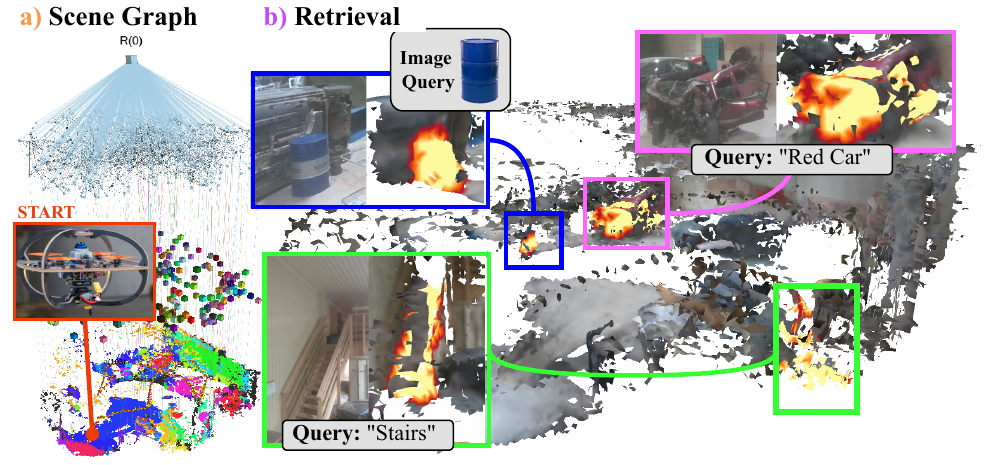}
    \caption{Aerial mapping: \textbf{a)} shows the open-vocabulary scene graph constructed online by \methodtitle~ from recorded aerial drone data, while \textbf{b)} demonstrates successful retrieval of all image- and text-queries.}
    \label{fig:drone-experiments}
    \vspace{-0.5em}
\end{figure}

\mypar{Aerial Mapping}
We further demonstrate \method{} on recorded drone RGB-D data from three indoor environments at a search-and-rescue training site, complementing standard ground-based benchmarks with uncommon aerial viewpoints. \cref{fig:drone-experiments} shows qualitative results from one environment.

% !TeX root = ../main.tex

\section{CONCLUSIONS}\label{sec:conclusions}

We present \method{}, an online open-vocabulary 3D scene graph method that maintains short-term mask identities through image-space DINOv3-based tracking, while using 3D association only to reconcile tracking interruptions and revisits. The resulting source tracks are fused into persistent 3D segments with compact multi-view open-vocabulary features. \method{} achieves competitive segmentation and retrieval compared with state-of-the-art open-vocabulary mapping methods while enabling efficient online operation with low memory use. Real-world experiments demonstrate onboard scene graph construction, retrieval, and object search on a quadruped robot, while recorded drone data demonstrates mapping and retrieval from aerial viewpoints.

\section*{ACKNOWLEDGMENT}
We thank the SYNERGISE project and S\"odert\"orns brandf\"orsvarsf\"orbund (SBFF) for providing access to their facilities for drone testing.

\bibliographystyle{IEEEtranN}
\bibliography{IEEEabrv,bib_short2}

% Generated by IEEEtranN.bst, version: 1.13 (2008/09/30)
\begin{thebibliography}{26}
\providecommand{\natexlab}[1]{#1}
\providecommand{\url}[1]{#1}
\csname url@samestyle\endcsname
\providecommand{\newblock}{\relax}
\providecommand{\bibinfo}[2]{#2}
\providecommand{\BIBentrySTDinterwordspacing}{\spaceskip=0pt\relax}
\providecommand{\BIBentryALTinterwordstretchfactor}{4}
\providecommand{\BIBentryALTinterwordspacing}{\spaceskip=\fontdimen2\font plus
\BIBentryALTinterwordstretchfactor\fontdimen3\font minus
  \fontdimen4\font\relax}
\providecommand{\BIBforeignlanguage}[2]{{%
\expandafter\ifx\csname l@#1\endcsname\relax
\typeout{** WARNING: IEEEtranN.bst: No hyphenation pattern has been}%
\typeout{** loaded for the language `#1'. Using the pattern for}%
\typeout{** the default language instead.}%
\else
\language=\csname l@#1\endcsname
\fi
#2}}
\providecommand{\BIBdecl}{\relax}
\BIBdecl

\bibitem[Hughes et~al.(2022)Hughes, Chang, et~al.]{hughes2022hydra}
N.~Hughes, Y.~Chang \emph{et~al.}, ``{Hydra}: A real-time spatial perception
  system for {3D} scene graph construction and optimization,'' \emph{Robotics:
  Science and Systems}, 2022.

\bibitem[Gu et~al.(2024)Gu, Kuwajerwala, et~al.]{Gu2024conceptgraphs}
Q.~Gu, A.~Kuwajerwala \emph{et~al.}, ``{ConceptGraphs}: Open-vocabulary {3D}
  scene graphs for perception and planning,'' \emph{IEEE International
  Conference on Robotics and Automation}, 2024.

\bibitem[Werby et~al.(2024)Werby, Huang, et~al.]{werby2023hovsg}
A.~Werby, C.~Huang \emph{et~al.}, ``Hierarchical open-vocabulary {3D} scene
  graphs for language-grounded robot navigation,'' \emph{Robotics: Science and
  Systems}, 2024.

\bibitem[Deng et~al.(2026)Deng, Tombari,
  et~al.]{deng_ovi-mapopen-vocabulary_2026}
Z.~Deng, F.~Tombari \emph{et~al.}, ``{OVI-MAP}: Open-vocabulary
  instance-semantic mapping,'' \emph{IEEE/CVF Conference on Computer Vision and
  Pattern Recognition}, 2026.

\bibitem[Laina et~al.(2026)Laina, Boche, et~al.]{laina_findanything_2026}
S.~B. Laina, S.~Boche \emph{et~al.}, ``{FindAnything}: Open-vocabulary and
  object-centric mapping for robot exploration in any environment,'' \emph{IEEE
  International Conference on Robotics and Automation}, 2026.

\bibitem[Zhao et~al.(2023)Zhao, Ding, et~al.]{zhao_fast_2023}
X.~Zhao, W.~Ding \emph{et~al.}, ``Fast segment anything,'' \emph{ArXiv}, 2023.

\bibitem[Radford et~al.(2021)Radford, Kim, et~al.]{Radford2021CLIP}
A.~Radford, J.~W. Kim \emph{et~al.}, ``Learning transferable visual models from
  natural language supervision,'' \emph{International Conference on Machine
  Learning}, 2021.

\bibitem[Sim{\'e}oni et~al.(2025)Sim{\'e}oni, Vo, et~al.]{simeoni_dinov3_2025}
O.~Sim{\'e}oni, H.~V. Vo \emph{et~al.}, ``{DINOv3},'' \emph{ArXiv}, 2025.

\bibitem[Kirillov et~al.(2023)Kirillov, Mintun, et~al.]{Kirillov2023SAM}
A.~Kirillov, E.~Mintun \emph{et~al.}, ``Segment anything,'' \emph{IEEE/CVF
  International Conference on Computer Vision}, 2023.

\bibitem[Peng et~al.(2023)Peng, Genova, et~al.]{Peng2023OpenScene}
S.~Peng, K.~Genova \emph{et~al.}, ``{OpenScene}: {3D} scene understanding with
  open vocabularies,'' \emph{IEEE/CVF Conference on Computer Vision and Pattern
  Recognition}, 2023.

\bibitem[Jatavallabhula et~al.(2023)Jatavallabhula, Kuwajerwala,
  et~al.]{Jatavallabhula2023conceptfusion}
K.~Jatavallabhula, A.~Kuwajerwala \emph{et~al.}, ``{ConceptFusion}: Open-set
  multimodal {3D} mapping,'' \emph{Robotics: Science and Systems}, 2023.

\bibitem[Zhou et~al.(2026)Zhou, Wei, et~al.]{zhou_opentrack3d_2025}
Z.~Zhou, S.~Wei \emph{et~al.}, ``{OpenTrack3D}: Towards accurate and
  generalizable open-vocabulary {3D} instance segmentation,'' \emph{IEEE/CVF
  Conference on Computer Vision and Pattern Recognition Findings Poster}, 2026.

\bibitem[Deng et~al.(2025)Deng, Yao, et~al.]{deng_openvox_2025}
Y.~Deng, B.~Yao \emph{et~al.}, ``{OpenVox}: Real-time instance-level
  open-vocabulary probabilistic voxel representation,'' \emph{IEEE/RSJ
  International Conference on Intelligent Robots and Systems}, 2025.

\bibitem[Bickici et~al.(2026)Bickici, Pabst, et~al.]{bickici2026thinkgraphs}
D.~Bickici, M.~Pabst \emph{et~al.}, ``Think while you map: Asynchronous
  vision-language agents for incremental {3D} scene graphs,'' \emph{European
  Conference on Computer Vision}, 2026.

\bibitem[Martins et~al.(2025)Martins, Oswald, et~al.]{martins2024ovoslam}
T.~B. Martins, M.~R. Oswald \emph{et~al.}, ``Open-vocabulary online semantic
  mapping for {SLAM},'' \emph{IEEE Robotics and Automation Letters}, 2025.

\bibitem[Yamazaki et~al.(2024)Yamazaki, Hanyu,
  et~al.]{yamazaki_open-fusion_2023}
K.~Yamazaki, T.~Hanyu \emph{et~al.}, ``{Open-Fusion}: Real-time open-vocabulary
  {3D} mapping and queryable scene representation,'' \emph{IEEE International
  Conference on Robotics and Automation}, 2024.

\bibitem[Kassab et~al.(2024)Kassab, Mattamala,
  et~al.]{kassab2024barenecessities}
C.~Kassab, M.~Mattamala \emph{et~al.}, ``The bare necessities: Designing
  simple, effective open-vocabulary scene graphs,'' \emph{ArXiv}, 2024.

\bibitem[Straub et~al.(2019)Straub, Whelan, et~al.]{replica19arxiv}
J.~Straub, T.~Whelan \emph{et~al.}, ``The {R}eplica dataset: A digital replica
  of indoor spaces,'' \emph{ArXiv}, 2019.

\bibitem[Yeshwanth et~al.(2023)Yeshwanth, Liu, et~al.]{yeshwanth2023scannet++}
C.~Yeshwanth, Y.-C. Liu \emph{et~al.}, ``Scannet++: A high-fidelity dataset of
  3d indoor scenes,'' \emph{IEEE/CVF International Conference on Computer
  Vision}, 2023.

\bibitem[Yadav et~al.(2023)Yadav, Ramrakhya, et~al.]{Yadav2023HM3DSem}
K.~Yadav, R.~Ramrakhya \emph{et~al.}, ``Habitat-matterport 3d semantics
  dataset,'' \emph{IEEE/CVF Conference on Computer Vision and Pattern
  Recognition}, 2023.

\bibitem[Kassab et~al.(2025)Kassab, Morin, et~al.]{kassab2025openlex3d}
C.~Kassab, S.~Morin \emph{et~al.}, ``{OpenLex3D}: A tiered benchmark for
  open-vocabulary {3D} scene representations,'' \emph{Conference on Neural
  Information Processing Systems}, 2025.

\bibitem[Schuhmann et~al.(2021)Schuhmann, Vencu,
  et~al.]{schuhmann_laion-400m_2021}
C.~Schuhmann, R.~Vencu \emph{et~al.}, ``{LAION}-{400M}: {Open} {Dataset} of
  {CLIP}-{Filtered} 400 {Million} {Image}-{Text} {Pairs},'' \emph{ArXiv}, 2021.

\bibitem[Zhai et~al.(2023)Zhai, Mustafa, et~al.]{Zhai2023Siglip}
X.~Zhai, B.~Mustafa \emph{et~al.}, ``Sigmoid loss for language image
  pre-training,'' \emph{IEEE/CVF International Conference on Computer Vision},
  2023.

\bibitem[Aharon et~al.(2022)Aharon, Orfaig, et~al.]{Aharon2022BoTSORTRA}
N.~Aharon, R.~Orfaig \emph{et~al.}, ``Bot-sort: Robust associations
  multi-pedestrian tracking,'' \emph{ArXiv}, 2022.

\bibitem[Dharmadhikari et~al.(2026)Dharmadhikari, Khedekar,
  et~al.]{Dharmadhikari2026uas}
M.~Dharmadhikari, N.~Khedekar \emph{et~al.}, ``The unified autonomy stack:
  Toward a blueprint for generalizable robot autonomy,'' \emph{ArXiv}, 2026.

\bibitem[Zacharia et~al.(2026)Zacharia, Dharmadhikari,
  et~al.]{zacharia2026omniplanner}
A.~Zacharia, M.~Dharmadhikari \emph{et~al.}, ``{OmniPlanner}: Universal
  exploration and inspection path planning across robot morphologies,''
  \emph{IEEE Transactions on Field Robotics}, 2026.

\end{thebibliography}

\end{document}